\documentclass[lettersize,journal]{IEEEtran}
\usepackage{amsmath,amsfonts}
\usepackage{algorithmic}
\usepackage{algorithm}
\usepackage{array}
\usepackage[caption=false,font=normalsize,labelfont=sf,textfont=sf]{subfig}
\usepackage{textcomp}
\usepackage{stfloats}
\usepackage{url}
\usepackage{verbatim}
\usepackage{graphicx}
\usepackage{cite}
\usepackage{xcolor}
\usepackage{graphicx}
\usepackage{xspace}
\usepackage{colortbl}
\usepackage{pifont}
\usepackage{multirow}
\usepackage{amsmath}
\usepackage{xparse}
\usepackage[extdef=true]{delimset}
\newcommand{\apr}{AP$_\textrm{r}$\xspace}
\newcommand{\apc}{AP$_\textrm{c}$\xspace}
\newcommand{\apf}{AP$_\textrm{f}$\xspace}
\newcommand{\apall}{AP$_\textrm{all}$\xspace}
\newcommand{\cmark}{\ding{51}}
\newcommand{\xmark}{\ding{55}}
\newcommand{\rotbox}[1]{\rotatebox{90}{#1}}

\newcommand{\M}{{\mathcal M}}
\newcommand{\xbf}{{\mathbf x}}
\newcommand{\ybf}{{\mathbf y}}
\newcommand{\prob}{p}
\newcommand{\smax}{\mathrm{softmax}}
\newcommand{\opus}[1]{%
  \begingroup
    \spaceskip=\fontdimen2\font plus \fontdimen3\font minus \fontdimen4\font
    \xspaceskip=\fontdimen7\font\relax
    \ttfamily
    #1%
  \endgroup
}
\definecolor{firstBest}{rgb}{0.9, 1, 0.9}
\definecolor{secondBest}{rgb}{1, 0.95, 0.95}
\definecolor{00blue}{RGB}{139,169,235}

\newcommand{\methodname}{RAR\xspace}
\newcommand{\hgreen}[1]{\textcolor{ForestGreen}{\textbf{#1}}} 
\newcommand{\hblue}[1]{\textcolor{NavyBlue}
{\textbf{#1}}} 

\definecolor{ForestGreen}{rgb}{0, 0.69, 0.31}
\definecolor{NavyBlue}{rgb}{0, 0.44, 0.75}

\definecolor{GrayBG}{gray}{0.95}

\newcommand{\yuhang}[1]{\textcolor{red}{[ToDo: #1]}}

\usepackage{booktabs}

\usepackage[numbers]{natbib}

\usepackage{hyperref}

\begin{document}

\title{RAR: Retrieving And Ranking Augmented \\
MLLMs for Visual Recognition}

\author{
Ziyu Liu$^{*1,2}$, Zeyi Sun$^{*1,2}$, Yuhang Zang$^{\dagger2}$, Wei Li$^{5}$, Pan Zhang$^{2}$, \\ Xiaoyi Dong$^{2}$, Yuanjun Xiong$^{4}$, Dahua Lin$^{2,3}$, Jiaqi Wang$^{\dagger2}$\\
$^1$Shanghai Jiao Tong University \quad 
$^2$Shanghai AI Laboratory \quad
$^3$The Chinese University of Hong Kong \quad 
$^4$MThreads, Inc.\quad  $^5$ Nanyang Technological University \\
}

\markboth{Journal of \LaTeX\ Class Files,~Vol.~14, No.~8, August~2021}%
{Shell \MakeLowercase{\textit{et al.}}: A Sample Article Using IEEEtran.cls for IEEE Journals}


\maketitle

\begin{abstract}
  CLIP (Contrastive Language–Image Pre-training) uses contrastive learning from noise image-text pairs to excel at recognizing a wide array of candidates, yet its focus on broad associations hinders the precision in distinguishing subtle differences among fine-grained items.
  Conversely, Multimodal Large Language Models (MLLMs) excel at classifying fine-grained categories, thanks to their substantial knowledge from pre-training on web-level corpora.
  However, the performance of MLLMs declines with an increase in category numbers, primarily due to growing complexity and constraints of limited context window size.
  To synergize the strengths of both approaches and enhance the few-shot/zero-shot recognition abilities for datasets characterized by extensive and fine-grained vocabularies, this paper introduces \methodname, a Retrieving And Ranking augmented method for MLLMs.
  We initially establish a multi-modal retriever based on CLIP to create and store explicit memory for different categories beyond the immediate context window.
  During inference, \methodname retrieves the top-$k$ similar results from the memory and uses MLLMs to rank and make the final predictions.
  Our proposed approach not only addresses the inherent limitations in fine-grained recognition but also preserves the model's comprehensive knowledge base, significantly boosting accuracy across a range of vision-language recognition tasks.
  Notably, our approach demonstrates a significant improvement in performance on 5 fine-grained visual recognition benchmarks, 11 few-shot image recognition datasets, and the 2 object detection datasets under the zero-shot recognition setting.
\end{abstract}



\section{Introduction}
\label{sec:intro}

\begin{figure*}[t]
  \centering
  \includegraphics[width=.95\linewidth]{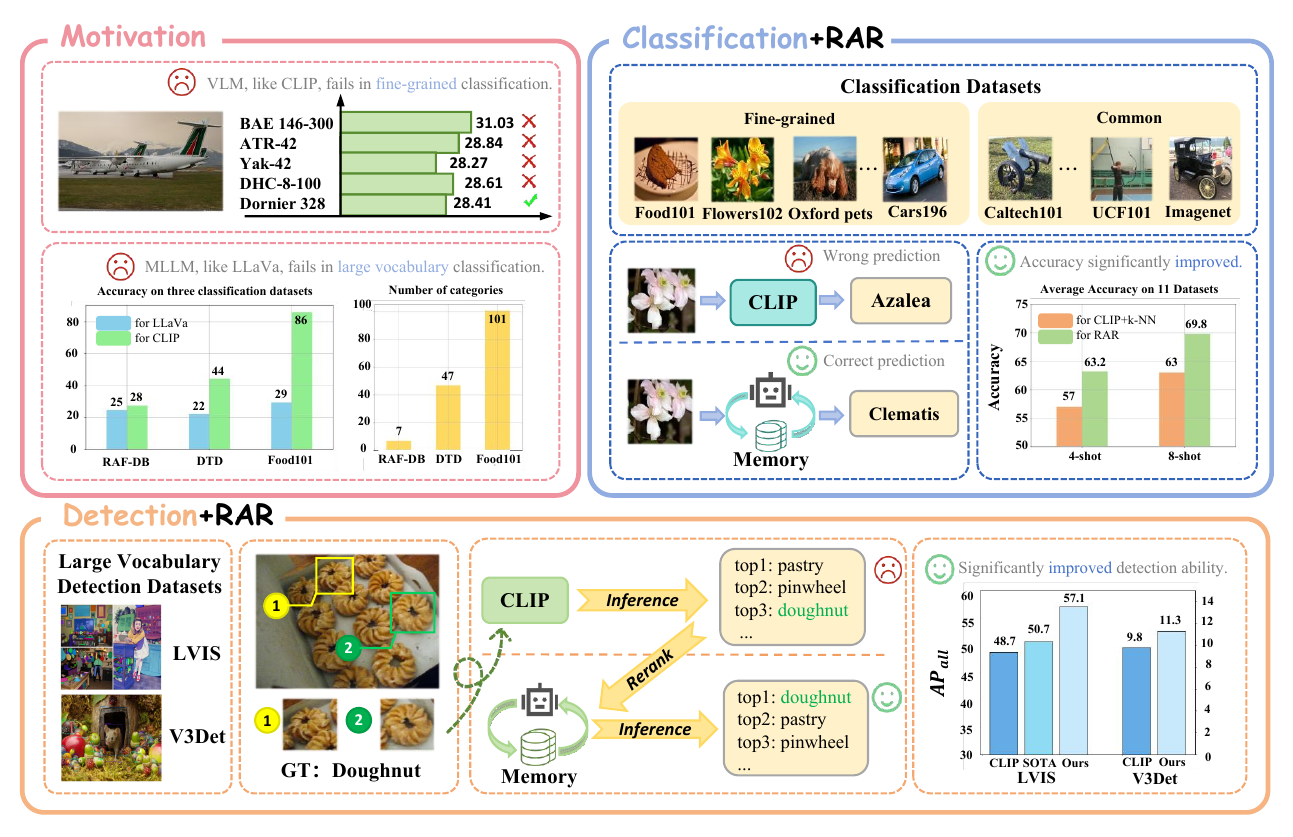}
  \vspace{-12pt}
  \caption{\textbf{Upper left}: our motivation about the drawbacks of CLIP and MLLM.
  Our \methodname can seamlessly integrate into MLLMs to improve the few-shot/zero-shot abilities on classification (\textbf{upper right}) and detection (\textbf{bottom}) datasets.}
  \label{fig:teaser}
  \vspace{-12pt}
\end{figure*}

\IEEEPARstart{T}{he} CLIP (Contrastive Language–Image Pre-training)~\citep{radford2021learning} model and its diverse variants \citep{EVA-CLIP,Dong_2023_CVPR,Li_2023_CVPR} provide flexible and robust performance across a wide array of visual-language understanding tasks.
Despite its successes, we observe that CLIP's performance begins to wane when faced with datasets characterized by vast vocabularies or fine-grained categories.
As shown in the upper left of Fig.~\ref{fig:teaser}, the decline is largely attributable to the inherent ambiguity of language descriptions and the challenges posed by synonyms, which can confound the model's ability to distinguish between closely related but distinct classes.

Parallel to these developments, Multi-modal Large Language Models (MLLMs) have emerged as a powerful class of generative models, exemplified by the likes of GPT-4V~\citep{2023gpt4vision} and analogous advancements~\citep{zhu2023minigpt,liu2024visual,dai2023instructblip,peng2023kosmos,ye2023mplug,awadalla2023openflamingo,zhang2023internlm,bai2023qwen,wang2023cogvlm,chen2023sharegpt4v, liu2023dynamic, liu2022hs}.
MLLMs, pre-trained on extensive corpora with substantial knowledge, demonstrate remarkable proficiency in identifying fine-grained categories when the total number of candidates remains manageable.
Nevertheless, MLLMs' efficacy is similarly compromised in scenarios involving extensive vocabularies and fine-grained categorizations (upper left of Fig.~\ref{fig:teaser}). 



To address these challenges, we propose augmenting standard MLLMs with our \methodname, a retrieving-and-ranking augmented technique.
Our \methodname enables models to dynamically incorporate external knowledge into the processing and generation workflows.
By augmenting MLLMs with external knowledge sources, we address challenges related to language ambiguity, synonym handling, and the limitations imposed by limited context windows when dealing with vast vocabularies.
Our method uses the inherent strength of MLLMs in generalizing from existing knowledge while addressing their limitations in visual recognition.
We first construct a multi-modal retriever that creates and stores multimodal embeddings for visual images and text descriptions.
As shown in Fig.\ref{fig:teaser}, upon receiving an input image at the inference stage, our approach retrieves the top-$k$ class names most similar to the image.
Subsequently, the MLLMs rank these retrieved candidate results as the final prediction results.
To bolster the MLLMs' ranking performance, we explore fine-tuning with ranking format data or in-context learning examples without training.
By integrating our retrieval-augmented design, our approach seeks to bridge the gap between the broad generalization capabilities of MLLMs and the need for precise, fine-grained categorization, offering a path forward that preserves the model's extensive knowledge base while significantly boosting its performance on downstream tasks.

To evaluate our method's efficacy, we conducted benchmarks in three areas: (1) fine-grained visual recognition across 5 benchmarks, (2) few-shot image recognition across 11 datasets, and (3) zero-shot object recognition on 2 object detection datasets with vast vocabularies (\textit{e.g.}, 13204 classes of V3Det~\citep{wang2023v3det}).
As presented in the right part of Fig.~\ref{fig:teaser}, our findings reveal that our approach notably enhances few-shot learning abilities, yielding an average improvement of $6.2\%$ over 11 image classification datasets under the 4-shot setting.
Furthermore, our method achieves a $6.4\%$ improvement on the LVIS dataset and a $1.5\%$ gain on the V3Det dataset in zero-shot object recognition performance.

In summary, our key contributions are outlined as follows: (1) We conduct an in-depth analysis of the strengths and weaknesses of VLMs and MLLMs in processing fine-grained datasets. (2) To enhance the fine-grained few-shot and zero-shot perception capabilities of MLLMs, we introduce \methodname with a multi-modal retriever and the inference pipeline based on retrieving and ranking. (3) Our \methodname can be seamlessly integrated into various MLLMs in a plug-and-play manner. (4) Through rigorous testing across 11 classification datasets and 2 object detection datasets, we demonstrate that our method outperforms baselines on a variety of visual recognition tasks.

\vspace{-6pt}

\section{Related Work}

\noindent \textbf{Contrastive Language-Image Pre-training (CLIP)}~\citep{radford2021learning} understands images and texts by contrastive learning from a vast amount of visual data paired with natural language descriptions. CLIP has robust capabilities in downstream tasks including image-text retrieval~\citep{yasunaga2022retrieval,Yu2024RankRAGUC,glass2022re2g}, zero-shot classification~\citep{zhou2021coop,gao2023clip}, and open-vocabulary perception~\citep{gu2021open,zang2022open,zhou2022detecting}.
Following CLIP, many subsequent vision-language models~\citep{jia2021scaling,li2022blip,li2022grounded,zhong2022regionclip,fang2023eva,Dong_2023_CVPR,Li_2023_CVPR,sun2023alpha,fgvp,clipseg} are proposed to further improve the vision-language understanding abilities. There are also works done to improve CLIP in zero-shot perception tasks~\citep{ReCLIP,circleCLIP,MaskAdaptedCLIP,maskQCLIP,yang2023recognize}. However, simple dot-product between two unimodality features can lead to sub-optimal results for fine-grained classification.
In this paper, we demonstrate that CLIP faces challenges in making accurate zero-shot predictions for fine-grained classes, and how our proposed method can effectively re-rank these predictions to improve the accuracy.

\noindent \textbf{Multimodal Large Language Models} (MLLMs) such as GPT4V~\citep{2023gpt4vision}, represent a significant evolution in the landscape of Large Language Models (LLMs) by integrating visual images as input tokens alongside textual information.
The integration is facilitated through the use of an additional vision encoder~\citep{radford2021learning} and a bridging mechanism~\citep{zhu2023minigpt,liu2024visual,dai2023instructblip,peng2023kosmos,ye2023mplug,awadalla2023openflamingo,zhang2023internlm,bai2023qwen,wang2023cogvlm,chen2023sharegpt4v}.
MLLMs significantly enhance the interaction between humans and AI in more natural and intuitive ways and demonstrate remarkable capabilities in understanding and generating multi-modal content.
Despite their prowess, our research uncovers a nuanced limitation: MLLMs tend to underperform in tasks requiring vast vocabularies, where distinguishing subtle differences among different categories is crucial. However, we prove that MLLMs exhibit a strong ability to excel in the re-ranking of top results obtained through vision-language models such as CLIP.
Fine-R~\citep{liu2024democratizing} first delves into leveraging MLLMs for fine-grained perception tasks by prompt design for better descriptions and attributes. We find a new way to prompt it with possible candidates to help screening and achieve better performance.
\section{Methodology}

\begin{figure*}[t]
  \centering
  \includegraphics[width=.95\linewidth]{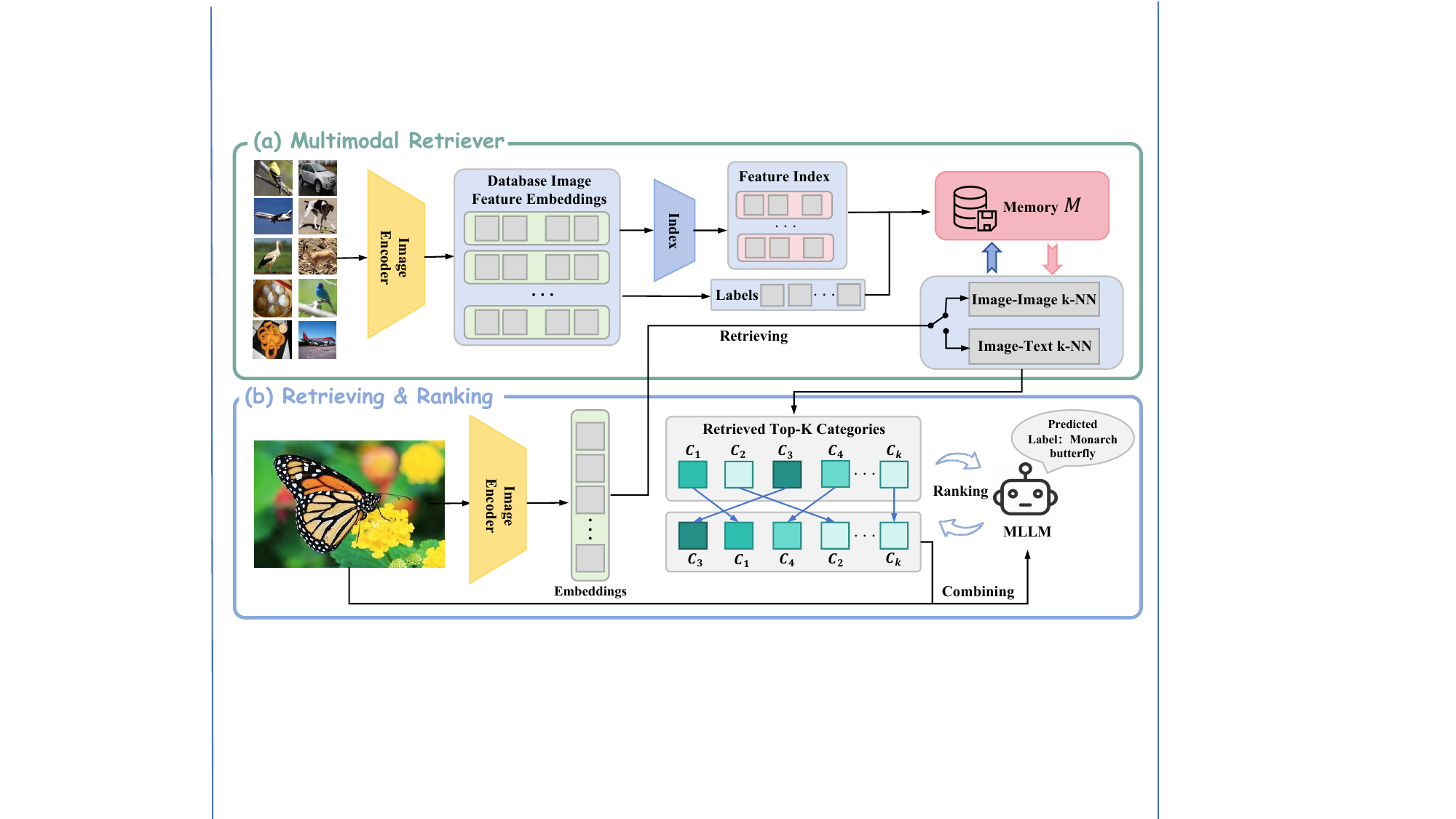}
  \vspace{-12pt}
  \caption{\textbf{Pipeline of \methodname.}
  \textbf{(a)} We design a \textbf{multimodal retriever} that extracts the image or text embeddings and stores embeddings in an external memory $\mathcal{M}$. \textbf{(b)} For the inference stage of downstream recognition tasks, we \textbf{retrieve} top-$k$ categories from the memory and use MLLMs to refine the retrieved results as the final prediction through \textbf{ranking}.}
  \label{fig:pipeline}
  \vspace{-12pt}
\end{figure*}

We first provide the background information on CLIP, MLLMs, and retrieval-augmentation in LLMs (Sec.~\ref{sec:preliminary}). Then we present the multi-modal retriever (Sec.~\ref{sec:retriever}) module of \methodname and how to apply \methodname on downstream tasks via retrieving and ranking (Sec.~\ref{sec:retrieving}).

\subsection{Preliminaries}\label{sec:preliminary}
\noindent \textbf{CLIP} is a model combining an image encoder $\Phi_{\text{img}}$ and a text encoder $\Phi_{\text{txt}}$ that uses contrastive learning to understand and align images and text by training on a vast dataset gathered from the web.
The core mechanism of CLIP involves mapping an input image $\mathcal{I}$ to its most semantically similar category $c \in \mathcal{C}$:
\vspace{-6pt}
\begin{equation}
    \prob \brk*{ y =c | \xbf } = \arg \max _{c \in \mathcal{C}}
    \cos(
    \Phi_{\text{img}}(\mathcal{I}), \Phi_{\text{txt}}(c)
    )
\text{\,,}
\vspace{-6pt}
\end{equation}

where $y$ represents the predicted category, $\mathcal{C}$ refers to the whole categories list and $\cos(\cdot,\cdot)$ denotes to the cosine similarity.

\noindent \textbf{Multimodal Large Language Models} such as GPT4V~\citep{2023gpt4vision} learning to generate predictions over sequences of tokens that span both image and text modalities.
The MLLM model $f$, parameterized by weights $\theta$, conditioned on the input sequences $\xbf = (x_1, \ldots, x_{L_{in}})$ of length $L_{in}$, which consist of both text tokens $\xbf_{\text{txt}}$ and visual tokens $\xbf_{\text{img}}$.
The $\xbf_{\text{img}}$ are extracted from the input image $\mathcal{I}$ via the image encoder $\Phi_{\text{img}}$.
MLLM model forecast a sequence of output tokens $\ybf = (y_1, \ldots, y_{L_{out}})$ of length $L_{out}$ as follows:
\begin{equation}
\begin{split}
    \prob_\theta \brk*{ \ybf | \xbf } = &\prod\nolimits_{l = 1}^{L_{out}} \prob_\theta \brk*{ y_l | \xbf, \ybf_{\leq l - 1} } \\
    = &\prod\nolimits_{l = 1}^{L_{out}} \smax \brk*{ f \brk*{ \xbf, \ybf_{\leq l - 1} ; \theta } }_{y_l}
\text{\,,}
\end{split}
\end{equation}
where $\ybf_{\leq l - 1}:= (y_1, \ldots, y_{l - 1})$ refers to the mechanism that predicts the distribution of the next token considering all previously generated tokens.

\noindent \textbf{Retrieval-Augmentation in Large Language Models} introduces a retrieval module $R$ with the LLM parameterized by $\theta$ for generation.
The retrieval module $R$ is designed to process an input sequence $\xbf$ against an external memory of documents $\M$, efficiently selecting a subset of documents $M\subseteq \M$.
The subset $M$ is then fed along with the original input sequence $\xbf$ into the LLM $\theta$, which uses both the input and the context provided by retrieved results to generate the target output $\ybf$:
\begin{equation}
    \prob_\theta \brk*{\ybf | \xbf, M} = \prod\nolimits_{l = 1}^{L_{out}} \prob_\theta \brk*{ y_l | \xbf, M, \ybf_{\leq l - 1} }.
\vspace{-6pt}
\end{equation}

\subsection{Multimodal Retriever}\label{sec:retriever}
The multimodal retriever is essentially responsible for querying a large multi-modal external memory or database to find information relevant to the input query or context.
In the process of multimodal retriever, the main challenge lies in efficiently encoding and storing a large volume of images/text embeddings for quick, accurate retrieval.
Recognizing the main challenge, as shown in Fig.~\ref{fig:pipeline}, we have developed a multi-modal retriever that creates and stores multimodal embeddings, with a focus on optimizing retrieval speed through index construction techniques.

\noindent \textbf{Extracting the Multi-modal Embeddings.}
We use the CLIP model discussed in Sec.~\ref{sec:preliminary} to extract the multi-modal embeddings.
Given a data sample $(x_{i}, c_{i})$ from the dataset $\mathcal{D}$ containing the image $x_{i}$ and class name $c_{i}$, we use the CLIP image encoder $\Phi_{\text{img}}$ to extract the image embedding $e_{\text{img}} \in \mathbb{R}^{d}$ and the CLIP text encoder $\Phi_{\text{text}}$ to extract the text embedding $e_{\text{text}} \in \mathbb{R}^{d}$.
The symbol $d$ refers to the feature dimension (\textit{e.g.}, $d=512$ for CLIP ViT-B/16).
The image and text embeddings are stored in the memory $\M$ for retrieval (will discuss in Sec.~\ref{sec:retrieving}).
In some zero-shot settings, the image embedding is not available and we merely store the text embedding into the memory. We build the Memory $\M$ (knowledge base) using these embeddings which are randomly sampled from the training set. In the $n$-shot setting, we collect $n$ images per class to construct the memory.

\noindent \textbf{Fast Retrieval Optimization.}
The brute force search is the common method for designing the retriever, which requires iteration over all vectors in the memory $\M$ to compute similarity scores (\textit{e.g.}, cosine similarity) and subsequently identify the top-$k$ results.
Although the brute force method is inherently straightforward, its efficiency markedly diminishes as the dataset escalates to the magnitude of millions of embeddings.
To enhance the speed of retrieval, we implement an index system that employs the HNSW (Hierarchical Navigable Small World) algorithm~\citep{malkov2018efficient}.
The HNSW methodology constructs an efficient index structure that enables approximate nearest neighbor search, thereby significantly accelerating the retrieval process without compromising much accuracy. This approximate search mechanism allows for rapid similarity retrieval in high-dimensional vector spaces while maintaining competitive precision compared with exact search methods.

\begin{figure*}[t]
  \centering
  \includegraphics[width=.95\linewidth]{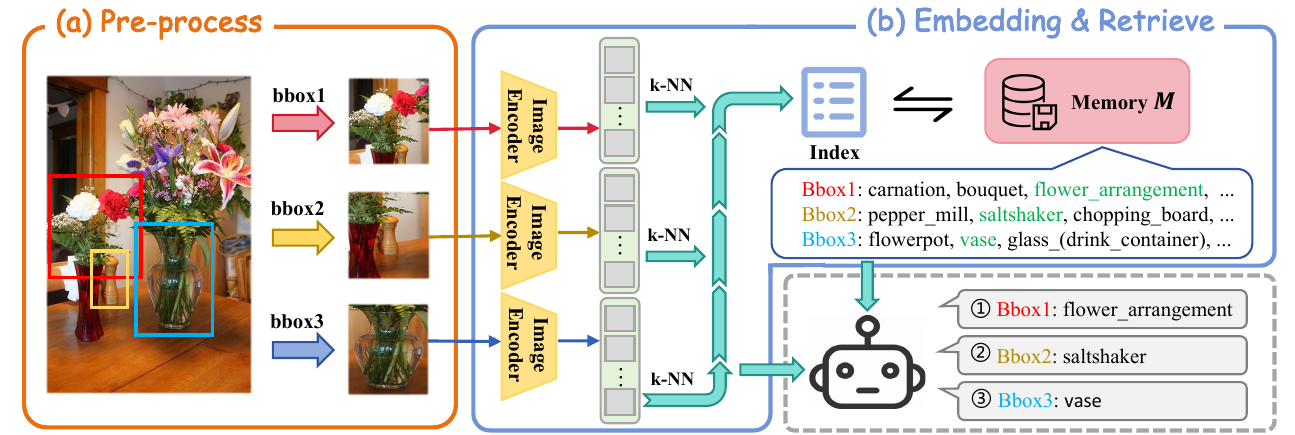}
  \caption{Extending our multimodal retriever to \textbf{zero-shot recognition} on object detection datasets such as LVIS~\citep{gupta2019lvis} and V3Det~\citep{wang2023v3det}. Compared to the classification datasets, we apply the additional pre-processing techniques such as \textbf{cropping} and \textbf{resizing} to extract the image embeddings.}
  \label{fig:pipeline_det}
  \vspace{-16pt}
\end{figure*}

\noindent \textbf{Pre-processing for Detection Datasets.} In object detection datasets, our methodology for extracting image embeddings $e_{\text{img}}$ is slightly different from the approach discussed previously.
As presented in Fig.~\ref{fig:pipeline_det}, we apply two additional pre-processing steps: cropping and blurring.
Some previous works have proposed similar methods in CLIP like \citep{fgvp,clipseg}.
In the object detection dataset, an image typically contains multiple objects of varying sizes.
Some objects may dominate a large portion of the image, whereas others occupy minimal space.
Accordingly, our object detection procedure begins with \textbf{cropping} the image regions based on proposal bounding box coordinates, subsequently \textbf{resizing} the cropped region to a fixed proportion.
Moreover, unlike image classification tasks the objects of interest generally appear large and centrally positioned, the objects within object detection datasets are smaller and their positions more varied.
To help the MLLMs understand the objects to be detected, we employ a \textbf{blurring} technique on the non-target areas surrounding the objects of interest.
The blurring strategy is designed to direct the MLLMs' focus toward the relevant objects, thereby facilitating their identification in object detection tasks.

\subsection{Inference with \textcolor{00blue}{R}etrieving \textcolor{00blue}{A}nd \textcolor{00blue}{R}anking}\label{sec:retrieving}
After successfully constructing memory $\M$ by using our multimodal retriever, our next step is to integrate the memory with the retrieval process and use MLLMs to rank the retrieval results and enhance the performance in few-shot/zero-shot perception tasks.

For example, in the inference stage of the few-shot image classification task, we first use the visual encoder $\Phi_{\text{img}}$ to process the input image $I$ and obtain the corresponding image embedding $\hat{e}$:
\begin{equation}
    \hat{e} = \Phi_{\text{img}}(I).
\end{equation}

The visual encoder is identical to the encoder used in our multi-modal retriever.
The image embedding $\hat{e}$ is then navigated through the previously constructed memory index and ranked by similarity to identify the top-$k$ related images.
Consequently, memory $\M$ yields the names of the retrieved top-$k$ categories, denoted as $\{c_1, c_2, c_3,..., c_k\}$:
\begin{equation}
\{ c_1, c_2, \dots, c_k \} = \text{Top-k}(\{\mathbf{e}_j\}_{j=1}^{N}, \hat{e}) \quad \mathbf{e}_j \in \mathcal{M},
\end{equation}
where $N$ is the number of embeddings stored in the memory $\mathcal{M}$.

The top-k retrieved results serve as a preliminary filter, narrowing down the vast possibilities to those most likely relevant, based on historical data and the semantic closeness of stored labels to the image content.

Since these cropped sub-images are usually small, CLIP's ability to extract features from these low-resolution images is limited. Therefore, in the object detection task, we do not perform image-to-image retrieval but use CLIP's inherent image-text interaction capabilities to conduct image-to-text retrieval.
Finally, we also obtain the top-$k$ category information with the highest similarity.

Following the retrieval phase, the retrieved category labels alongside image embedding $\hat{e}$ are integrated and sent to the MLLMs through our ranking prompt.
The MLLMs, combining the internal knowledge and the retrieved information, make the final prediction of the image category. This process can be formalized as:
\begin{equation}
    \hat{y} = \text{Rank}(\text{Top-k}(\{\mathbf{e}_j\}_{j=1}^{N}, \hat{e}), P_{\text{rank}}) \quad \mathbf{e}_j \in \mathcal{M},
\end{equation}
where $\hat{y}$ denotes the final prediction result, and $P_{\text{rank}}$ represents the prompt used for ranking.

Our proposed inference process, using both the retrieval results from our memory bank and subsequent ranking by the MLLM, ensures a more accurate and contextually aware classification prediction.
Our design represents a significant advancement in few-shot image classification, enabling our system to handle a wide variety of images and categories with high precision and flexibility.

\begin{figure*}[t]
  \centering
  \includegraphics[width=.95\linewidth]{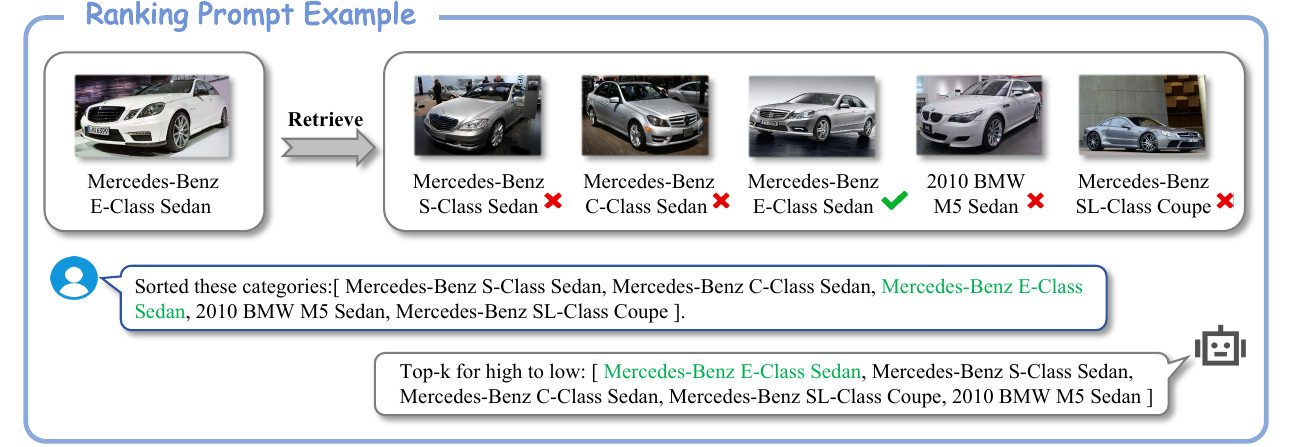}
  \vspace{-6pt}
  \caption{\textbf{Ranking Prompt examples} for few-shot image classification. The fine-grained image examples are from Stanford Cars~\citep{krause20133d}.
  We incorporate the initial top-$k$ retrieved results (\textit{e.g.}, $k=5$) into our ranking prompts and use the MLLMs to rank the retrieved results and make the final prediction.
  }
  \label{fig:prompt}
  \vspace{-12pt}
\end{figure*}

\noindent \textbf{Ranking Prompt Format.} Fig.~\ref{fig:prompt} presents an example ranking prompt format. The process begins with the prompt \texttt{`Sort the optional categories: [class a, class b, class c, class d, class e]'}, which is dynamically generated to include the top-k class names retrieved from our multimodal retriever.
Our method uses the MLLM's ability to rank these retrieved class names. Unlike traditional approaches that might rely solely on the initial retrieval order, our MLLM employs advanced linguistic and semantic analysis to assess the contextual appropriateness of each class name with the input image. 

\noindent \textbf{Fine-tuning for Ranking.}
When directly applying MLLMs to ranking the retrieved results, MLLMs may predict some errors such as beyond the given list or occasional misalignment.
To fully exploit the ranking potential of MLLMs for downstream tasks, while avoiding the consumption of extensive computational resources for training MLLMs, we selected a small-scale classification dataset to fine-tune the MLLMs.
The primary goal of fine-tuning was to enable MLLMs to improve their ranking ability such as following the format of prompts and returning results as required.

To create our fine-tuning data, we use the CLIP image encoder $\Phi_{\text{img}}$ to extract the embeddings of two disjoint subsets of images $\mathcal{D}_{a}$ and $\mathcal{D}_{b}$, both drawn from the FGVC-Aircraft dataset.
We provide the ablation studies in Sec.~\ref{sec:exp_ablation} about using different datasets to construct the fine-tuning data.
Our observation reveals that the MLLM demonstrates robustness to the choice of fine-tuning datasets, with only marginal differences in performance outcomes.

For each image in $\mathcal{D}_{b}$, we apply the $k$-NN clustering algorithm to find the top 20 most similar images in $\mathcal{D}_{a}$ including their categories.
Afterward, we select 16 sets from these 20 images, each set comprising $k$ images, and retain those groups that contain images of the same category as $\mathcal{D}_{b}$.
We then shuffled the category labels for these sets.
Using similar prompts shown in Fig.~\ref{fig:prompt}, we create a dataset comprising roughly 30,000 entries, with the original sequence of categories serving as the ground-truth label.
In summary, we build the fine-tuning data aiming to bolster the MLLM's ranking performance.

\noindent \textbf{In-Context Learning for Ranking.}
In-context learning presents a valuable alternative to fine-tuning with ranking examples, particularly due to its flexibility and lower requirement for specialized data preparation.
While fine-tuning with ranking examples has proven to be highly effective, it necessitates a substantial amount of curated data and computational resources for training.
In contrast, in-context learning uses the model's existing knowledge by providing it with specific examples directly within the input prompt, guiding the model to understand and execute the task of ranking without the need for explicit re-training.
Here we elaborate on the application of in-context learning with MLLMs to rank the retrieved results.
To effectively guide the MLLMs in comprehending the ranking task, we use the prompt format similar to Fig.~\ref{fig:prompt} and integrate a specific ranking example into the prompts.
Please refer to the Sec.~\ref{sec:exp_setup} for our structured in-context learning prompt.
Please refer to Sec.~\ref{sec:exp_ablation} for the ablation studies of discussing the difference between using fine-tuning or in-context learning for ranking.
\section{Experiments}

\begin{figure}[t]
  \centering
  \includegraphics[width=1.\linewidth]{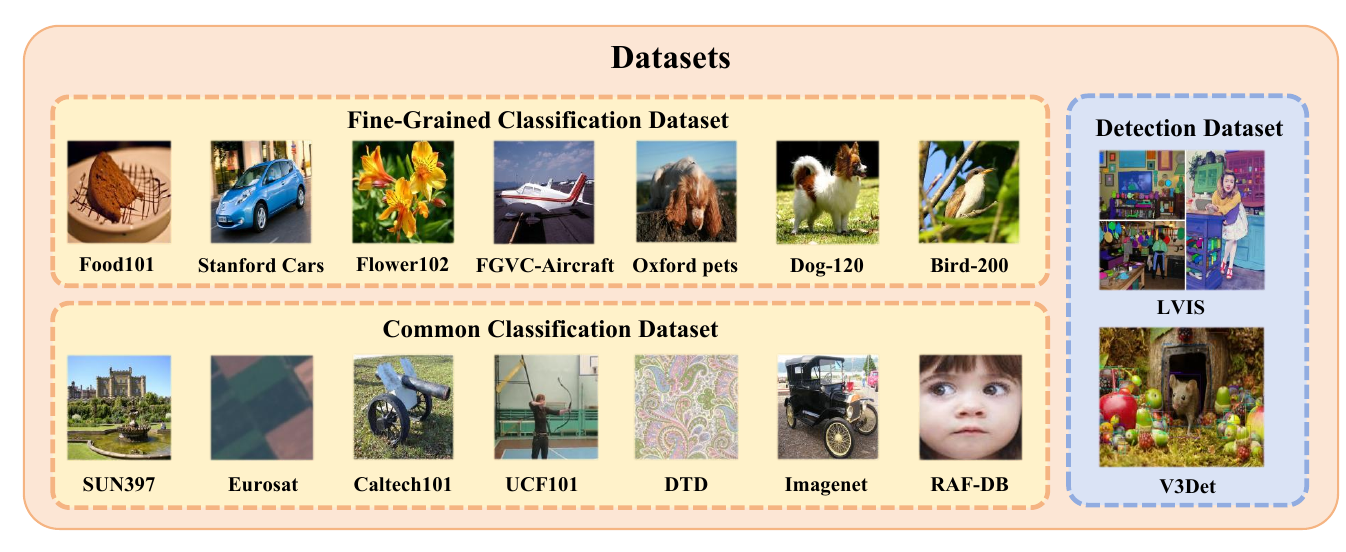}
  \vspace{-16pt}
  \caption{\textbf{Datsets} used in our experiments. We select 14 classification datasets (7 fine-grained and 7 common) and 2 object detection datasets as our benchmarks.}
  \label{fig:datasets}
  \vspace{-12pt}
\end{figure}

\begin{table}[t]
  \caption{
  Statistics for the classification and detection datasets used in our three settings: fine-grained visual recognition, few-shot image recognition, and zero-shot region recognition.
  }
  \label{tab:dataset_statistic}
  \centering
  \renewcommand{\arraystretch}{0.5}
    \resizebox{1.\linewidth}{!}{
  \begin{tabular}{@{}c|ccccc@{}}
  \toprule
  Settings & Dataset & Categories & Evaluation Metrics & Source link \\

  \cmidrule(r){1-1} \cmidrule(r){2-2} \cmidrule(r){3-5}
  
   \multirow{5}{*}{ \parbox{2.3cm}{Fine-Grained Visual Recog.}}  & Bird-200 & 200 & cACC, sACC & \href{https://data.caltech.edu/records/65de6-vp158}{Bird website}\\
  ~ & Car-196 & 196 & cACC, sACC & \href{https://www.kaggle.com/datasets/jessicali9530/stanford-cars-dataset}{Kaggle}\\
  ~ & Dog-120 & 120 & cACC, sACC & \href{https://www.tensorflow.org/datasets/catalog/stanford_dogs}{Tensorflow}\\
  ~ & Flower-102 & 102 & cACC, sACC & \href{https://www.tensorflow.org/datasets/catalog/oxford_flowers102}{Tensorflow}\\
  ~ & Pet-37 & 37 & cACC, sACC & \href{https://www.tensorflow.org/datasets/catalog/oxford_iiit_pet}{Tensorflow}\\

   \cmidrule(r){1-1} \cmidrule(r){2-2} \cmidrule(r){3-5}

   \multirow{12}{*}{\parbox{2.3cm}{Few-Shot Image Recog.}} & RAF-DB & 7 & Accuracy & \href{http://www.whdeng.cn/raf/model1.html}{RAF-DB website}\\
   ~ & Eurosat & 10 & Accuracy & \href{https://www.tensorflow.org/datasets/catalog/eurosat}{Tensorflow}\\
   ~ & DTD & 47 & Accuracy & \href{https://www.tensorflow.org/datasets/catalog/dtd}{Tensorflow}\\
   ~ & FGVC Aircraft & 100 & Accuracy & \href{https://www.robots.ox.ac.uk/~vgg/data/fgvc-aircraft/}{FGVC website}\\
   ~ & Caltech101 & 101 & Accuracy & \href{https://www.tensorflow.org/datasets/catalog/caltech101}{Tensorflow}\\
   ~ & Food101 & 101 & Accuracy & \href{https://www.tensorflow.org/datasets/catalog/food101}{Tensorflow}\\
   ~ & UCF-101 & 101 &Accuracy & \href{https://www.tensorflow.org/datasets/catalog/ucf101}{Tensorflow}\\
   ~ & SUN397 & 397 & Accuracy & \href{https://www.tensorflow.org/datasets/catalog/sun397}{Tensorflow}\\
   ~ & ImageNet & 1000 & Accuracy & \href{https://www.tensorflow.org/datasets/catalog/imagenet2012}{Tensorflow}\\
   \cmidrule(r){1-1} \cmidrule(r){2-2} \cmidrule(r){3-5}

   Zero-Shot     & LVIS &  1203 & mAP &\href{https://www.lvisdataset.org/}{LVIS website}\\
   Region Recog. & V3Det & 13204 & mAP & \href{https://github.com/V3Det}{Github}\\

  \bottomrule
  \end{tabular}}
  \vspace{-12pt}
\end{table}

\begin{table*}[!t]
    \vspace{-18pt}
    \small
    \begin{center}
    \caption{
    Fine-grained visual recognition across 5 datasets. We follow~\citep{liu2024democratizing} to report the averaged clustering accuracy (cACC, \%) and semantic similarity accuracy (sACC, \%) results.
    The best and second-best results are colored \colorbox{firstBest}{\bf Green} and \colorbox{secondBest}{Red}, respectively.
    }
    \label{tab:exps_fgvr}
    \vspace{-12pt}
    \renewcommand{\arraystretch}{0.5}
    \resizebox{1.\linewidth}{!}{
    \begin{tabular}{@{}l@{\ \ \ \ }|c@{\ \ }c@{\ \ }c@{\ \ }c@{\ \ }c@{\ \ }c@{\ \ }c@{\ \ }c@{\ \ }c@{\ \ }c@{\ \ }|c@{\ \ }c@{\ \ }}
    \toprule
     & \multicolumn{2}{c}{Bird-200\xspace} & \multicolumn{2}{c}{Car-196\xspace} & \multicolumn{2}{c}{Dog-120\xspace} & \multicolumn{2}{c}{Flower-102\xspace} & \multicolumn{2}{c|}{Pet-37\xspace} & \multicolumn{2}{c}{Average} \\
    & cACC & sACC & cACC & sACC & cACC & sACC & cACC & sACC & cACC & sACC & cACC & sACC \\
    \cmidrule(r){1-1} \cmidrule(r){2-3} \cmidrule(r){4-5} \cmidrule(r){6-7} \cmidrule(r){8-9} \cmidrule(r){10-11} \cmidrule(r){12-13}
    WordNet+CLIP
        & 39.3 & 57.7
        & 18.3 & 33.3
        & \colorbox{firstBest}{\bf 53.9} & \colorbox{firstBest}{\bf 70.6}
        & 42.1 & 49.8
        & 55.4 & 61.9
        & 41.8 & 54.7 \\
        
    BLIP-2
        & 30.9 & 56.8
        & 43.1& 57.9
        & 39.0 & 58.6
        & 61.9 & \colorbox{firstBest}{\bf 59.1}
        & 61.3 & 60.5
        & 47.2 & 58.6 \\

        
        
    CaSED
        & 25.6 & 50.1
        & 26.9 & 41.4
        & 38.0 & 55.9
        & \colorbox{firstBest}{\bf 67.2} & 52.3
        & 60.9 & 63.6
        & 43.7 & 52.6 \\

    FineR
        & 51.1 & \colorbox{firstBest}{\bf 69.5}
        & \colorbox{secondBest}{49.2} & \colorbox{secondBest}{ 63.5}
        &  48.1 &  64.9
        & \colorbox{secondBest}{ 63.8 } &  51.3
        & \colorbox{secondBest}{72.9} 
        & \colorbox{secondBest}{72.4}
        & \colorbox{secondBest}{57.0}
        & \colorbox{secondBest}{64.3} \\

    \cmidrule(r){1-1} \cmidrule(r){2-3} \cmidrule(r){4-5} \cmidrule(r){6-7} \cmidrule(r){8-9} \cmidrule(r){10-11} \cmidrule(r){12-13}

    \methodname(Ours) & \colorbox{firstBest}{\bf 51.6}
    & \colorbox{firstBest}{\bf 69.5}
    & \colorbox{firstBest}{\bf 53.2}
    & \colorbox{firstBest}{\bf 63.6}
    & \colorbox{secondBest}{50.0} 
    & \colorbox{secondBest}{65.2} & 63.7 
    & \colorbox{secondBest}{53.2}
    & \colorbox{firstBest}{\bf 74.1 }
    & \colorbox{firstBest}{\bf 74.8 }
    & \colorbox{firstBest}{\bf 58.5 }
    & \colorbox{firstBest}{\bf 65.3} \\
    \bottomrule     
    \end{tabular}}
    \end{center}
    \vspace{-12pt}
\end{table*}

\begin{table*}[t]
  \setlength{\tabcolsep}{1mm}
  \footnotesize
  \caption{
  \footnotesize
  Few-shot image classification across 11 datasets. We report the top-1 accuracy (\%) under the 1-shot, 2-shot, 4-shot, 8-shot and 16-shot settings. The CLIP+KNN method does not utilize the text encoder of CLIP. Instead, we employ the visual encoder to extract image features, and then apply the KNN algorithm to these features. Here our \methodname uses the LLaVA1.5~\citep{liu2023improved} as the MLLM to rank the retrieved results. The symbol `-' denotes to the LLaVA model fails to make the predictions due to the limited window size.
  }
  \label{tab:few_shot_appendix}
  \vspace{-6pt}
  \renewcommand{\arraystretch}{0.5}
  \centering
  \resizebox{1.\linewidth}{!}{
  \begin{tabular}{l |cccccc ccccc| c}
  \toprule
  Method & \multicolumn{7}{c}{Common} 
  & \multicolumn{4}{c|}{Fine-Grained} \\
  \cmidrule(r){1-1} \cmidrule(lr){2-8} \cmidrule(lr){9-12} \\
  & \rotbox{ImageNet} & \rotbox{Caltech101} & \rotbox{RAF-DB} & \rotbox{SUN397} & \rotbox{EuroSAT} & \rotbox{DTD} & \rotbox{UCF-101}
  & \rotbox{Flower102} & \rotbox{StanfordCars} & \rotbox{Food101} & \rotbox{OxfordPets} & \rotbox{\textbf{Average}} \\
  \cmidrule(r){1-1} \cmidrule(r){2-8} \cmidrule(r){9-12} \cmidrule(r){13-13}
  
  \texttt{1-shot} \\
  CLIP+KNN & \colorbox{secondBest}{29.2} & 75.9 & 11.3 & \colorbox{secondBest}{37.7} & \colorbox{secondBest}{53.9} & \colorbox{secondBest}{35.1} & \colorbox{secondBest}{47.8} & \colorbox{firstBest}{\bf66.7} & \colorbox{secondBest}{32.6} & \colorbox{secondBest}{45.3} & \colorbox{secondBest}{41.3} & \colorbox{secondBest}{43.3} \\
  LLaVA1.5 Finetuning & - & \colorbox{secondBest}{84.1} & \colorbox{secondBest}{24.9} & - & 48.2 & 22.3 & 35.4  & 4.59 & - & 39.2 & 16.3 & - \\
  \methodname (LLaVA1.5) &\colorbox{firstBest}{\bf 40.3} & \colorbox{firstBest}{\bf85.2} & \colorbox{firstBest}{\bf34.8} & \colorbox{firstBest}{\bf 46.5}  &  \colorbox{firstBest}{\bf62.4} & \colorbox{firstBest}{\bf38.1} & \colorbox{firstBest}{\bf57.4} &  \colorbox{secondBest}{50.4} & \colorbox{firstBest}{\bf38.3} & \colorbox{firstBest}{\bf57.6} & \colorbox{firstBest}{\bf47.0} & \colorbox{firstBest}{\bf50.7}  \\
  $\Delta$ & \hgreen{+10.5} & \hgreen{+9.3} & \hgreen{+23.5} & \hgreen{+8.8} & \hgreen{+8.5} & \hgreen{+3.0} & \hgreen{+9.6} &  \hblue{-16.3} & \hgreen{+5.7} & \hgreen{+12.3} & \hgreen{+5.7} & \hgreen{+7.4} \\
  \cmidrule(r){1-1} \cmidrule(r){2-8} \cmidrule(r){9-12} \cmidrule(r){13-13}

  \texttt{2-shot} \\
  CLIP+KNN & \colorbox{secondBest}{36.1} & \colorbox{secondBest}{82.9} & 11.7 &\colorbox{secondBest}{ 44.6} & \colorbox{secondBest}{58.7} & \colorbox{secondBest}{41.2} & \colorbox{secondBest}{58.5} & \colorbox{firstBest}{\bf78.9} & \colorbox{secondBest}{40.9} & \colorbox{secondBest}{54.1} & \colorbox{secondBest}{49.0} & \colorbox{secondBest}{50.6} \\
  LLaVA1.5 Finetuning & - & 53.1 & \colorbox{secondBest}{24.9} & - & 48.2 & 22.3 & 38.7 & 10.03  & - & 38.2 & 16.3 & -\\
  \methodname (LLaVA1.5) & \colorbox{firstBest}{\bf46.8} & \colorbox{firstBest}{\bf89.2} & \colorbox{firstBest}{\bf27.9} & \colorbox{firstBest}{\bf53.1} & \colorbox{firstBest}{\bf68.6} & \colorbox{firstBest}{\bf47.9} & \colorbox{firstBest}{\bf66.5} & \colorbox{secondBest}{54.7}  & \colorbox{firstBest}{\bf45.9} & \colorbox{firstBest}{\bf65.4} &  \colorbox{firstBest}{\bf54.7} & \colorbox{firstBest}{\bf57.4}\\
  $\Delta$ & \hgreen{+10.7} & \hgreen{+6.3} & \hgreen{+16.2} & \hgreen{+8.5} & \hgreen{+9.9} & \hgreen{+6.7} & \hgreen{+8.0} & \hblue{-24.2}  & \hgreen{+5.0} &\hgreen{+11.3}  & \hgreen{+5.7} &  \hgreen{+6.8} \\
  \cmidrule(r){1-1} \cmidrule(r){2-8} \cmidrule(r){9-12} \cmidrule(r){13-13}

  \texttt{4-shot} \\
  CLIP+KNN & \colorbox{secondBest}{42.1} & 87.9 & 14.2 & \colorbox{secondBest}{51.4} & \colorbox{secondBest}{67.6} & \colorbox{secondBest}{47.5} & \colorbox{secondBest}{64.6} &  \colorbox{firstBest}{\bf 84.5} & \colorbox{secondBest}{49.2} & 62.6 & \colorbox{secondBest}{55.6} & \colorbox{secondBest}{57.0} \\
  LLaVA1.5 Finetuning & - & \colorbox{secondBest}{88.4} & \colorbox{secondBest}{24.9} &- & 48.2 & 46.6 & 58.9 & 13.2 & -& \colorbox{secondBest}{66.4} & 28.9 & -\\
  \methodname (LLaVA1.5) & \colorbox{firstBest}{\bf 51.0} & \colorbox{firstBest}{\bf 92.1} & \colorbox{firstBest}{\bf 27.7} & \colorbox{firstBest}{\bf 58.8} & \colorbox{firstBest}{\bf 74.8} & \colorbox{firstBest}{\bf 53.9} & \colorbox{firstBest}{\bf 69.6} & \colorbox{secondBest}{80.4} & \colorbox{firstBest}{\bf 54.4} & \colorbox{firstBest}{\bf 71.4} & \colorbox{firstBest}{\bf 60.9} & \colorbox{firstBest}{\bf 63.2} \\
  $\Delta$ & \hgreen{+9.9} & \hgreen{+4.2} & \hgreen{+13.5} &\hgreen{+7.4} & \hgreen{+7.2}& \hgreen{+6.4}& \hgreen{+5.0} & \hblue{-4.1} &\hgreen{+5.2}& \hgreen{+8.8}&\hgreen{+5.3}&\hgreen{+6.2}\\
  \cmidrule(r){1-1} \cmidrule(r){2-8} \cmidrule(r){9-12} \cmidrule(r){13-13}

  \texttt{8-shot} \\
  CLIP+KNN & \colorbox{secondBest}{47.6} & 90.6 & \colorbox{secondBest}{28.2} & \colorbox{secondBest}{56.8} & \colorbox{secondBest}{72.8} & 53.2 & \colorbox{secondBest}{68.3} & \colorbox{firstBest}{\bf 89.5} & \colorbox{secondBest}{56.1} & 68.3 & \colorbox{secondBest}{61.8} & \colorbox{secondBest}{63.0} \\
  LLaVA1.5 Finetuning & -& \colorbox{secondBest}{92.1} & 24.9 & -& 48.2 & \colorbox{secondBest}{54.7} & 66.5 & 30.1 &- & \colorbox{secondBest}{72.5} & 46.1& -\\
  \methodname (LLaVA1.5) & \colorbox{firstBest}{\bf 56.5} & \colorbox{firstBest}{\bf 93.5} & \colorbox{firstBest}{\bf 46.9} & \colorbox{firstBest}{\bf 63.4} & \colorbox{firstBest}{\bf 81.5} & \colorbox{firstBest}{\bf 59.3} & \colorbox{firstBest}{\bf 74.3} & \colorbox{secondBest}{87.3} & \colorbox{firstBest}{\bf 61.2} & \colorbox{firstBest}{\bf 76.6} & \colorbox{firstBest}{\bf 67.7} & \colorbox{firstBest}{\bf 69.8} \\
  $\Delta$&\hgreen{+8.9}&\hgreen{+2.9}&\hgreen{+18.7}&\hgreen{+6.6}&\hgreen{+8.7}&\hgreen{+6.1}
  &\hgreen{+6.0}&\hblue{-2.2}&\hgreen{+5.1}&\hgreen{+8.3}&\hgreen{+5.9}&\hgreen{+6.8}\\
  \cmidrule(r){1-1} \cmidrule(r){2-8} \cmidrule(r){9-12} \cmidrule(r){13-13}

  \texttt{16-shot} \\
  CLIP+KNN & \colorbox{secondBest}{52.0} & 92.4 & \colorbox{secondBest}{35.0} & \colorbox{secondBest}{61.2} & \colorbox{secondBest}{78.7} & 57.5 & 70.6 & \colorbox{secondBest}{92.1} & \colorbox{secondBest}{63.2} & \colorbox{secondBest}{71.8} & \colorbox{secondBest}{68.3} & \colorbox{secondBest}{67.5} \\
  LLaVA1.5 Finetuning & - & \colorbox{secondBest}{94.1} & 24.9 & - & 50.6 & \colorbox{secondBest}{63} & \colorbox{secondBest}{74.7} & 59.0 & - & - & 62.4 & - \\
  \methodname (LLaVA1.5) & \colorbox{firstBest}{\bf60.3} & \colorbox{firstBest}{\bf94.1} &  \colorbox{firstBest}{\bf53.1} & \colorbox{firstBest}{\bf68.0}  & \colorbox{firstBest}{\bf84.8} & \colorbox{firstBest}{\bf63.7} & \colorbox{firstBest}{\bf75.9} & \colorbox{firstBest}{\bf92.1}  & \colorbox{firstBest}{\bf67.8} & \colorbox{firstBest}{\bf79.4} & \colorbox{firstBest}{\bf72.7}& \colorbox{firstBest}{\bf73.8}  \\
  $\Delta$ & \hgreen{+8.3} & \hgreen{+1.7} & \hgreen{+18.1} & \hgreen{+6.8} & \hgreen{+6.1} & \hgreen{+6.2} & \hgreen{+5.3} &  \hgreen{+0.0} & \hgreen{+4.6} & \hgreen{+7.6} & \hgreen{+4.4} & \hgreen{+6.3} \\
  \bottomrule
  \end{tabular}
  \vspace{-12pt}
  }
\end{table*}

\begin{table}[t]
  \setlength{\tabcolsep}{1mm}
  \footnotesize
  \caption{
  \footnotesize
  Evaluation on 11 datasets, reporting the top-1 accuracy. We use the CLIP ViT-L/14@336 as feature extractor and \methodname is based on LLaVA 1.5.
  }
  \label{tab:clip_v14_classification}
  \renewcommand{\arraystretch}{0.5}
  \centering
  \resizebox{1.\linewidth}{!}{
  \begin{tabular}{l |cccccc ccccc| c}
  \toprule
  Method & \multicolumn{7}{c}{Common} 
  & \multicolumn{4}{c|}{Fine-Grained} \\
  \cmidrule(r){1-1} \cmidrule(lr){2-8} \cmidrule(lr){9-12} \\
  & \rotbox{ImageNet} & \rotbox{Caltech101} & \rotbox{RAF-DB} & \rotbox{SUN397} & \rotbox{EuroSAT} & \rotbox{DTD} & \rotbox{UCF-101}
  & \rotbox{Flower102} & \rotbox{StanfordCars} & \rotbox{Food101} & \rotbox{OxfordPets} & \rotbox{\textbf{Average}} \\
  \cmidrule(r){1-1} \cmidrule(r){2-8} \cmidrule(r){9-12} \cmidrule(r){13-13}

  \texttt{4-shot} \\
  CLIP+KNN & \colorbox{secondBest}{52.2} & \colorbox{secondBest}{92.4} & \colorbox{secondBest}{24.7} & \colorbox{secondBest}{56.2} & \colorbox{secondBest}{68.3} & \colorbox{secondBest}{52.5} & \colorbox{secondBest}{72.6} & \colorbox{firstBest}{\bf 92.3} & \colorbox{secondBest}{62.4} & \colorbox{secondBest}{74.1} & \colorbox{secondBest}{67.0} & \colorbox{secondBest}{65.0} \\
  \methodname (LLaVA1.5) & \colorbox{firstBest}{\bf58.4} & \colorbox{firstBest}{\bf93.6} & \colorbox{firstBest}{\bf46.3} & \colorbox{firstBest}{\bf61.8} & \colorbox{firstBest}{\bf73.5} & \colorbox{firstBest}{\bf58.5} & \colorbox{firstBest}{\bf75.5} & \colorbox{secondBest}{83.2} & \colorbox{firstBest}{\bf70.8} & \colorbox{firstBest}{\bf79.0} & \colorbox{firstBest}{\bf68.4} & \colorbox{firstBest}{\bf69.9}\\
  $\Delta$ & \hgreen{+6.2} & \hgreen{+1.2} & \hgreen{+21.6} & \hgreen{+5.6} & \hgreen{+5.2}& \hgreen{+6.0} & \hgreen{+2.9} &  \hblue{-9.1} & \hgreen{+8.4} & \hgreen{+4.9} & \hgreen{+1.4} & \hgreen{+4.9}\\
  \cmidrule(r){1-1} \cmidrule(r){2-8} \cmidrule(r){9-12} \cmidrule(r){13-13}

  \texttt{8-shot} \\
  CLIP+KNN & \colorbox{secondBest}{57.8} & \colorbox{secondBest}{94.4} & \colorbox{secondBest}{41.0} & \colorbox{secondBest}{61.3} & \colorbox{secondBest}{78.9} & \colorbox{secondBest}{57.0} & \colorbox{secondBest}{76.2} & \colorbox{firstBest}{\bf95.8} & \colorbox{secondBest}{63.1} & \colorbox{secondBest}{80.2} & \colorbox{secondBest}{73.1} & \colorbox{secondBest}{70.8} \\
 \methodname (LLaVA1.5) & \colorbox{firstBest}{\bf63.2} & \colorbox{firstBest}{\bf95.0} & \colorbox{firstBest}{\bf57.6} & \colorbox{firstBest}{\bf66.9} & \colorbox{firstBest}{\bf84.3} & \colorbox{firstBest}{\bf62.8} & \colorbox{firstBest}{\bf79.1} & \colorbox{secondBest}{93.2}
 & \colorbox{firstBest}{\bf70.8} & \colorbox{firstBest}{\bf83.5} & \colorbox{firstBest}{\bf73.7} & \colorbox{firstBest}{\bf75.5}\\
  $\Delta$& \hgreen{+5.4} & \hgreen{+0.6} & \hgreen{+16.6} & \hgreen{+5.6} & \hgreen{+5.4} & \hgreen{+5.8} & \hgreen{+2.9} & \hblue{-2.6} & \hgreen{+7.7} & \hgreen{+3.3} & \hgreen{+0.6} & \hgreen{+4.7}\\
  \bottomrule
  \end{tabular}
  \vspace{-6pt}
  }
\end{table}

\begin{table}[t]
  \setlength{\tabcolsep}{1mm}
  \footnotesize
  \caption{
  \footnotesize
  Evaluation on 11 datasets, reporting the top-5 accuracy. We use the 4-shot setting.
  }
  \label{tab:top5}
  \vspace{-6pt}
  \renewcommand{\arraystretch}{0.5}
  \centering
  \resizebox{1.\linewidth}{!}{
  \begin{tabular}{l |cccccc ccccc| c}
  \toprule
  Method & \multicolumn{7}{c}{Common} 
  & \multicolumn{4}{c|}{Fine-Grained} \\
  \cmidrule(r){1-1} \cmidrule(lr){2-8} \cmidrule(lr){9-12} \\
  & \rotbox{ImageNet} & \rotbox{Caltech101} & \rotbox{RAF-DB} & \rotbox{SUN397} & \rotbox{EuroSAT} & \rotbox{DTD} & \rotbox{UCF-101}
  & \rotbox{Flower102} & \rotbox{StanfordCars} & \rotbox{Food101} & \rotbox{OxfordPets} & \rotbox{\textbf{Average}} \\
  \cmidrule(r){1-1} \cmidrule(r){2-8} \cmidrule(r){9-12} \cmidrule(r){13-13}
  
  CLIP+KNN & \colorbox{secondBest}{67.1} & \colorbox{secondBest}{97.6} & \colorbox{secondBest}{48.0} & \colorbox{secondBest}{78.9} & \colorbox{secondBest}{91.5} & \colorbox{secondBest}{70.5} & \colorbox{secondBest}{85.4} & \colorbox{secondBest}{96.5} & \colorbox{secondBest}{79.1} & \colorbox{secondBest}{86.2} & \colorbox{secondBest}{87.6} & \colorbox{secondBest}{80.8}\\
 \cmidrule(r){1-1} \cmidrule(r){2-8} \cmidrule(r){9-12} \cmidrule(r){13-13}
  
  \methodname (LLaVA1.5) & \colorbox{firstBest}{\bf69.7} & \colorbox{firstBest}{\bf97.7} & \colorbox{firstBest}{\bf53.8} & \colorbox{firstBest}{\bf80.1} & \colorbox{firstBest}{\bf92.5} & \colorbox{firstBest}{\bf71.9} & \colorbox{firstBest}{\bf86.2} & \colorbox{firstBest}{\bf96.5} & \colorbox{firstBest}{\bf79.1} & \colorbox{firstBest}{\bf87.7} & \colorbox{firstBest}{\bf88.1}& \colorbox{firstBest}{\bf82.1} \\
  $\Delta$ &\hgreen{+2.6} & \hgreen{+0.1}& \hgreen{+5.8}& \hgreen{+1.2}& \hgreen{+1.0}& \hgreen{+1.4}& \hgreen{+0.8}& \hgreen{+0.0}& \hgreen{+0.0}& \hgreen{+1.5}& \hgreen{+0.5}& \hgreen{+1.3}\\

  \bottomrule
  \end{tabular}
  \vspace{-6pt}
  }
\end{table}

\begin{table}[t]
  \setlength{\tabcolsep}{1mm}
  \footnotesize
  \caption{
  \footnotesize
  Evaluation on 11 datasets, reporting the top-1 accuracy.
  The GPT4V~\citep{2023gpt4vision} results are copied from~\citep{wu2023gpt4vis}.
  }
  \label{tab:4v_classification}
  \renewcommand{\arraystretch}{0.5}
  \centering
  \resizebox{1.\linewidth}{!}{
  \begin{tabular}{l |cccccc ccccc| c}
  \toprule
  Method & \multicolumn{7}{c}{Common} 
  & \multicolumn{4}{c|}{Fine-Grained} \\
  \cmidrule(r){1-1} \cmidrule(lr){2-8} \cmidrule(lr){9-12} \\
  & \rotbox{ImageNet} & \rotbox{Caltech101} & \rotbox{RAF-DB} & \rotbox{SUN397} & \rotbox{EuroSAT} & \rotbox{DTD} & \rotbox{UCF-101}
  & \rotbox{Flower102} & \rotbox{StanfordCars} & \rotbox{Food101} & \rotbox{OxfordPets} & \rotbox{\textbf{Average}} \\
  \cmidrule(r){1-1} \cmidrule(r){2-8} \cmidrule(r){9-12} \cmidrule(r){13-13}
  
  GPT-4V & 62.0 & \colorbox{secondBest}{95.5} & 58.5 & 57.7 & 36.2 & 59.1 & \colorbox{firstBest}{\bf81.6} & 70.6 & 58.3 & 80.1 & \colorbox{firstBest}{\bf92.6} & 68.4 \\
 \cmidrule(r){1-1} \cmidrule(r){2-8} \cmidrule(r){9-12} \cmidrule(r){13-13}
  
  \methodname (LLaVA1.5) & \colorbox{secondBest}{73.4} & 94.6 & \colorbox{firstBest}{\bf73.8} & \colorbox{secondBest}{70.6} & \colorbox{firstBest}{\bf93.3} & \colorbox{secondBest}{71.9} & 79.1 & \colorbox{secondBest}{95.6} & \colorbox{secondBest}{72.6} & \colorbox{secondBest}{86.2} & 79.9 & \colorbox{secondBest}{81.0}   \\
  $\Delta$ &\hgreen{+11.4}&\hblue{-0.9}&\hgreen{+15.3}&\hgreen{+12.9}&\hgreen{+57.1}&\hgreen{+12.8}&\hblue{-2.5}&\hgreen{+25.0}&\hgreen{+14.3}&\hgreen{+6.1}&\hblue{-12.7}&\hgreen{+12.6}\\
  \cmidrule(r){1-1} \cmidrule(r){2-8} \cmidrule(r){9-12} \cmidrule(r){13-13}

  \methodname (Intern-IXC2) & 71.5 & 94.4 &\colorbox{secondBest}{ 72.7 }& 69.7 & \colorbox{secondBest}{91.7} & 69.9 & 77.6 & 93.2 & 65.4 & 83.9 & 79.3 &79.0 \\
  $\Delta$ &\hgreen{+9.5}&\hblue{-1.1}&\hgreen{+14.2}&\hgreen{+12.0}&\hgreen{+55.5}&\hgreen{+10.8}&\hblue{-4.0}&\hgreen{+22.6}&\hgreen{+7.1}&\hgreen{+3.8}&\hblue{-13.3}&\hgreen{+10.6}\\
  \cmidrule(r){1-1} \cmidrule(r){2-8} \cmidrule(r){9-12} \cmidrule(r){13-13}

    \methodname (Qwen-VL) & \colorbox{firstBest}{\bf75.8} & \colorbox{firstBest}{\bf95.5} & 66.0 & \colorbox{firstBest}{\bf72.7} & 90.7 & \colorbox{firstBest}{\bf72.5} & \colorbox{secondBest}{81.4} & \colorbox{firstBest}{\bf97.5} & \colorbox{firstBest}{\bf81.6} & \colorbox{firstBest}{\bf87.2} &\colorbox{secondBest}{88.1} & \colorbox{firstBest}{\bf82.6}\\
  $\Delta$ &\hgreen{+13.8}&\hgreen{+0.0}&\hgreen{+7.5}&\hgreen{+5.0}&\hgreen{+54.5}&\hgreen{+13.4}&\hblue{-0.2}&\hgreen{+26.9}&\hgreen{+23.3}&\hgreen{+7.1}&\hblue{-4.5}&\hgreen{+14.2}\\
  \bottomrule
  \end{tabular}
  \vspace{-6pt}
  }
\end{table}

In this section, we present our experiment step (Sec.~\ref{sec:exp_setup}) and conduct experiments on different tasks such as fine-grained visual recognition (Sec.~\ref{sec:fine-grained}), few-shot image recognition (Sec.~\ref{sec:exp_few_cls}) and zero-shot object recognition (Sec.~\ref{sec:exp_zero_det}). We also provide the ablation studies about our design choices (Sec.~\ref{sec:exp_ablation}).

\subsection{Experimental Setup}\label{sec:exp_setup}
\noindent \textbf{Datasets and Evaluation Metrics.}
We follow previous work~\citep{liu2024democratizing} to choose 5 datasets for \textbf{fine-grained visual recognition} (Bird-200~\citep{wahcub200}, Cars-196~\citep{krause20133d}, Dog-120~\citep{khosla2011novel}, Flower-102~\citep{nilsback2008automated}, and Pet-37~\citep{parkhi2012cats}) and report the clustering accuracy (cACC) and semantic similarity accuracy (sACC) as evaluation metrics.

For \textbf{few-shot image recognition}, we select 11 datasets including general objects (ImageNet~\citep{deng2009imagenet}, Caltech101~\citep{fei2004learning}), textual (DTD~\citep{cimpoi2014describing}), scene objects (SUN397~\citep{xiao2010sun}), satellite images (EuroSAT~\citep{helber2019eurosat}), facial expressions (RAF-DB~\citep{li2017reliable}), car types (Stanford Cars~\citep{krause20133d}) and fine-grained datasets (FGVC-Aircraft~\citep{maji2013fine}, Oxford Flowers~\citep{nilsback2008automated}, Food101~\citep{nilsback2008automated} and Oxford Pets~\citep{parkhi2012cats}).
We report the top-1 accuracy (\%) for all these classification datasets.

Additionally, we also select two benchmarks for our \textbf{zero-shot object recognition} setting: (1) The LVIS\citep{gupta2019lvis} dataset that encompasses over 164,000 images and 1,203 categories.
LVIS is characterized by its long-tailed distribution of objects, reflecting a broad spectrum of categories with varying degrees of rarity—from commonly seen items to those that are seldom encountered in everyday scenes.
We report the \apr, \apc, \apf, and \apall metrics for rare, common, frequent, and all categories.
(2) V3Det~\citep{wang2023v3det} dataset encompasses an immense number of 13204 categories of real-world images.
V3Det challenges current models to navigate and interpret an unprecedented scale of category diversity, driving the development of more sophisticated and nuanced detection algorithms.
For V3Det, we report the standard mAP metric of the object detection task.



We present all the utilized datasets in Fig.~\ref{fig:datasets}. And in Tab.~\ref{tab:dataset_statistic}, we list the statistics and sources of these datasets in detail.

\noindent \textbf{Prompt Formats}
In this section, we delve into the detailed design of our prompts. We have crafted distinct prompts for various tasks to test the capabilities of the baseline model and our \methodname model in visual recognition.

In our \methodname pipeline, the prompt primarily serves to merge the input image with the category information retrieved from memory. It guides MLLMs to rank the retrieved candidate object categories based on similarity. Our prompt format is as follows:

\opus{Please play the role of a classification expert, and sort the provided categories from high to low according to the \{top-k\} similarity with the input image. Here are the optional categories:\{categories\}.} 
 
Here, `\{top-k\}' is replaced with the number of categories input.  And `\{categories\}' is replaced with the top-k categories retrieved from memory.

Additionally, to assess the visual recognition and ranking capabilities of MLLMs themselves, we have prepared a prompt with examples to serve as input for the model. Our structured in-context learning prompt is as follows:

\opus{Please play the role of a classification expert, and sort the provided categories from high to low according to the top 5 similarity with the input image. Here are the optional categories:\{categories\}. Your answer should follow the following format, like:[`category A', `category B', `category C', `category D', `category E']. Only choose five categories, and no further information.} 

When testing the \methodname pipeline with MLLMs, ‘\{categories\}’ is replaced with all the category names of each dataset.

\noindent \textbf{Implementation Details.} We employ a frozen CLIP ViT B/16 model as the visual encoder $\Phi_{\text{img}}$ to encode the input images and extract the corresponding image embeddings.
For the retrieval process, we search the stored embeddings in memory $\M$ using the HNSW algorithm~\citep{malkov2018efficient}.
We use $k=5$ for the top-$k$ results, with a solo exception $k=4$ in the $4$-shot few-shot setting.
To improve the ranking ability of MLLMs, we prepare $30$k fine-tuning data from the FGVC-Aircraft dataset.
In the fine-tuning process, we train the model with one epoch with a learning rate of $1e^{-5}$ on our fine-tuning data and subsequently evaluate the performance across additional datasets.
We present the ablation studies about the hyper-parameters such as the value of $k$ and the fine-tuning data source in the Sec.~\ref{sec:exp_ablation}.

\subsection{Fine-Grained Visual Recognition}\label{sec:fine-grained}

We first evaluate our \methodname on the \textit{fine-grained visual recognition} setting defined in previous work ~\citep{liu2024democratizing}.
We use only 3 unlabelled images per category to build our memory $\M$ for retrieving.
We use two synergistic metrics: Clustering Accuracy (cACC) and Semantic Similarity (sACC) to evaluate our method, following~\citep{liu2024democratizing}. Clustering Accuracy (cACC) mainly assesses the accuracy of clustering images within the same category, without considering the semantic relatedness of category labels. Complementing this, Semantic Similarity (sACC) measures the similarity between the names of categories in the clusters and the ground truth.

\noindent \textbf{Baselines.} We follow~\citep{liu2024democratizing} to select four representative methods as our baselines to compare with: WordNet~\citep{miller1995wordnet}+CLIP, BLIP-2~\citep{li2023blip}, CaSED~\citep{conti2024vocabulary}, and FineR~\citep{liu2024democratizing}.

\noindent \textbf{Averaged Results over 5 Datasets.} Tab.~\ref{tab:exps_fgvr} summarizes the results and our \methodname achieves the top performance on both the cACC (58.5\%) and sACC (65.3\%) metrics.
The WordNet+CLIP and CaSED baselines rely solely on CLIP for class name retrieval, yet often yield inaccurate predictions.
In contrast, our method adds the additional ranking process with MLLMs, which increases the likelihood of correctly predicting those accurate yet initially lower-ranked candidates and thereby boosts performance.
Besides, FineR uses MLLM (\textit{e.g.}, BLIP-2) for fine-grained recognition via multi-round questioning-answering processes, which may demand more computational resources and struggle to scale efficiently with large vocabulary datasets.
Conversely, our approach first retrieves candidates and then lets MLLMs make predictions on the candidates, optimizing both accuracy and efficiency.

We can observe that \methodname did not achieve SOTA results on Dog-120 and Flower-102. This is because some baselines use exhaustive knowledge bases on specialized datasets: WordNet covers all ground-truth Dog-120 categories, and CaSED includes 101 of 102 ground-truth Flower-102 categories. As discussed in FineR~\citep{liu2024democratizing}, this leads to biased high performance. Moreover, BLIP-2 uses a more powerful 11B Flan-T5xxl encoder. \methodname does not use these exhaustive knowledge bases but still achieves the best results on majority fine-grained datasets (average +16.7\%/+11.3\%/+14.8\% gains over WordNet/BLIP-2/CaSED), which demonstrate \methodname is effective and general.

\subsection{Few-Shot Image Recognition}\label{sec:exp_few_cls}

The few-shot setting aims to enable a model to recognize new objects with only a few examples for each new category.
Few-shot learning faces substantial challenges when applied to fine-grained datasets, which consist of numerous highly similar classes yet are accompanied by only a minimal amount of training data.

\noindent \textbf{Baselines.}
For \textit{few-shot image recognition}, we introduce two baselines including CLIP and MLLMs.
The first is the CLIP~\citep{radford2021learning} model combined with $k$-NN to retrieve predictions based on few-shot examples.
The second is the LLaVA model directly fine-tuning with LoRA~\citep{hu2021lora} on few-shot examples.

\noindent \textbf{Averaged Results on 11 Datasets.} Tab.~\ref{tab:few_shot_appendix} summarizes the few-shot results on 11 datasets, including 4 fine-grained datasets.
Compared to the CLIP initial retrieval results (top row), our \methodname (third row) with ranking facilitates a notable increase in classification accuracy.
On average, our approach boosts the top-1 accuracy from 57.0 to 63.2 ($\%$) on the 4-shot setting (increased by 6.2$\%$), and from 63.0 to 69.8 ($\%$) on the 8-shot setting (increased by 6.8$\%$).
Such improvements illustrate the ranking process of MLLMs effectively uses a nuanced understanding of context and detail to better align predictions with ground truth.
Additionally, we observe that LLaVA1.5 + fine-tuning (second row) baseline underperforms in datasets with large vocabularies such as ImageNet due to the constraint of LLMs' context window.
Thanks to the retrieved candidates, our \methodname works for datasets with a vast of categories and is a potent tool in refining classification decisions, proving particularly useful in handling the diverse and challenging landscape of image classification tasks.

\noindent \textbf{More Feature Extractor results.}
In the few-shot image classification experiment, we primarily used the CLIP-B/16. To verify the generalization capability of the method, we conducted experiments using the more powerful CLIP ViT-L/14@336 as the feature extractor. The results, as shown in Tab.~\ref{tab:clip_v14_classification}, demonstrate that \methodname still consistently outperforms the baseline (increased by 4.9$\%$ on 4-shot setting, and 4.7$\%$ on 8-shot setting).

\noindent \textbf{Top-5 Accuracy Results.} Moreover, in the experiments conducted for our paper, we selected the top 5 retrieved results for ranking. To test the scalability of this method, we conducted a new experiment using the top 10 retrieved results, ranking these ten categories and then assessing the accuracy of the top 5. In this experiment, we utilized a 4-shot setting, the result is shown in Tab.~\ref{tab:top5}.

The final results demonstrate that although the top 5 accuracy achieved by CLIP+KNN was already high, our \methodname method still managed to make comprehensive improvements on this basis. The average top 5 accuracy across eleven datasets increased by 1.3\%.

\noindent \textbf{Extension to the whole Training Set.} To further explore the potential of \methodname, we expanded the memory size to include all images from the training set stored in memory. We then compared the performance of \methodname under this setup with that of GPT-4V across multiple image classification datasets. The results are presented in Tab.~\ref{tab:4v_classification}.

The results in Tab.~\ref{tab:4v_classification} show that, regardless of whether the base model is LLaVa, Intern-IXC2, or Qwen-VL, \methodname significantly outperforms GPT-4V in terms of accuracy. Across eleven datasets, the average precision of \methodname exceeds that of GPT-4V by 12.6$\%$. It is observed that even 7B MLLMs, when integrated into the \methodname pipeline, far surpass the classification capabilities of GPT-4V across multiple image classification datasets.

\begin{figure}[t]
  \centering
  \includegraphics[width=1.\linewidth]{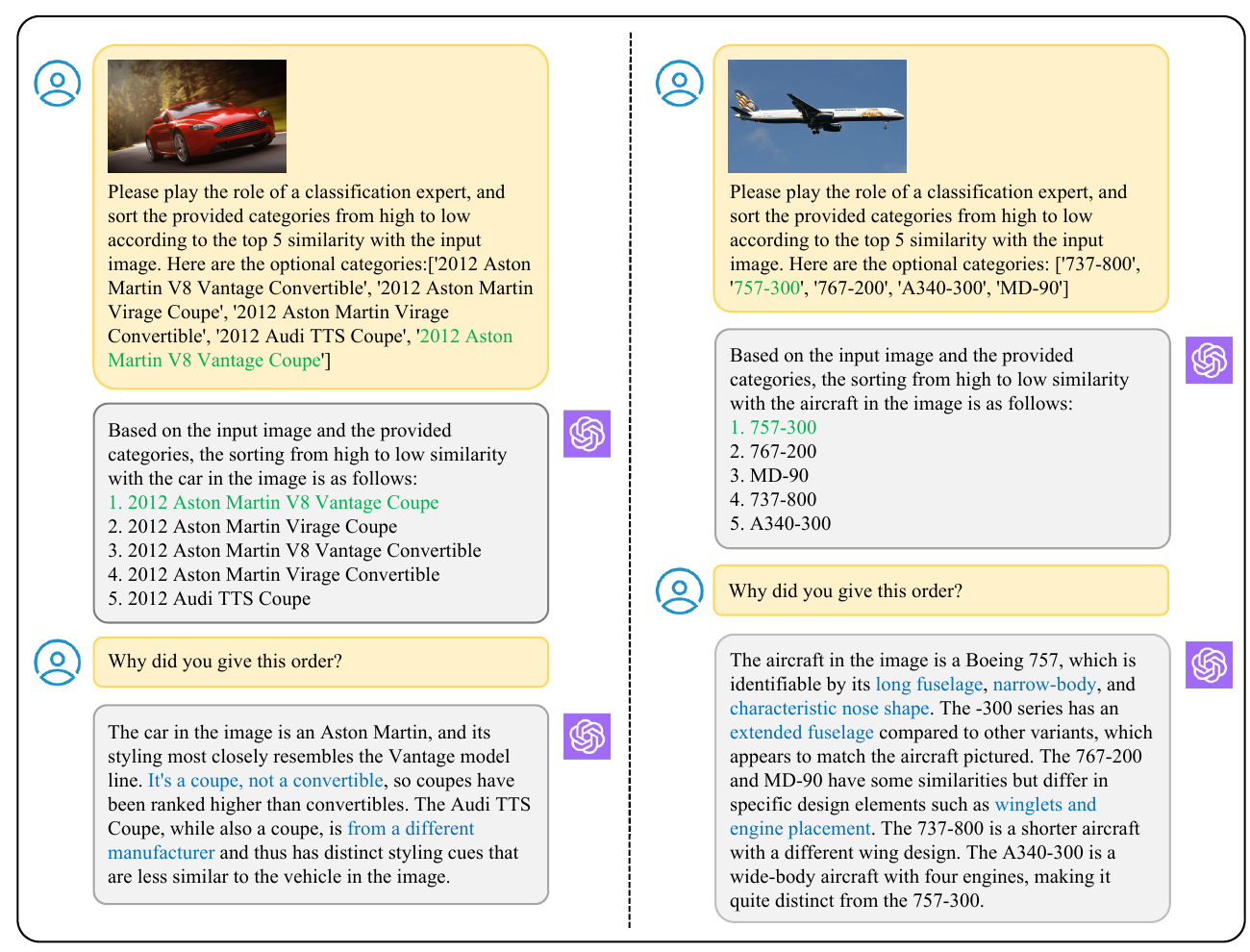}
  \vspace{-16pt}
  \caption{\textbf{GPT4V Example} for Stanford Cars and FGVC Aircraft. \hgreen{Green} for ground truth, \hblue{blue} for characteristics analyzed by GPT-4V. }
  \label{fig:4v1}
  \vspace{-12pt}
\end{figure}

\begin{figure}[t]
  \centering
  \includegraphics[width=1.\linewidth]{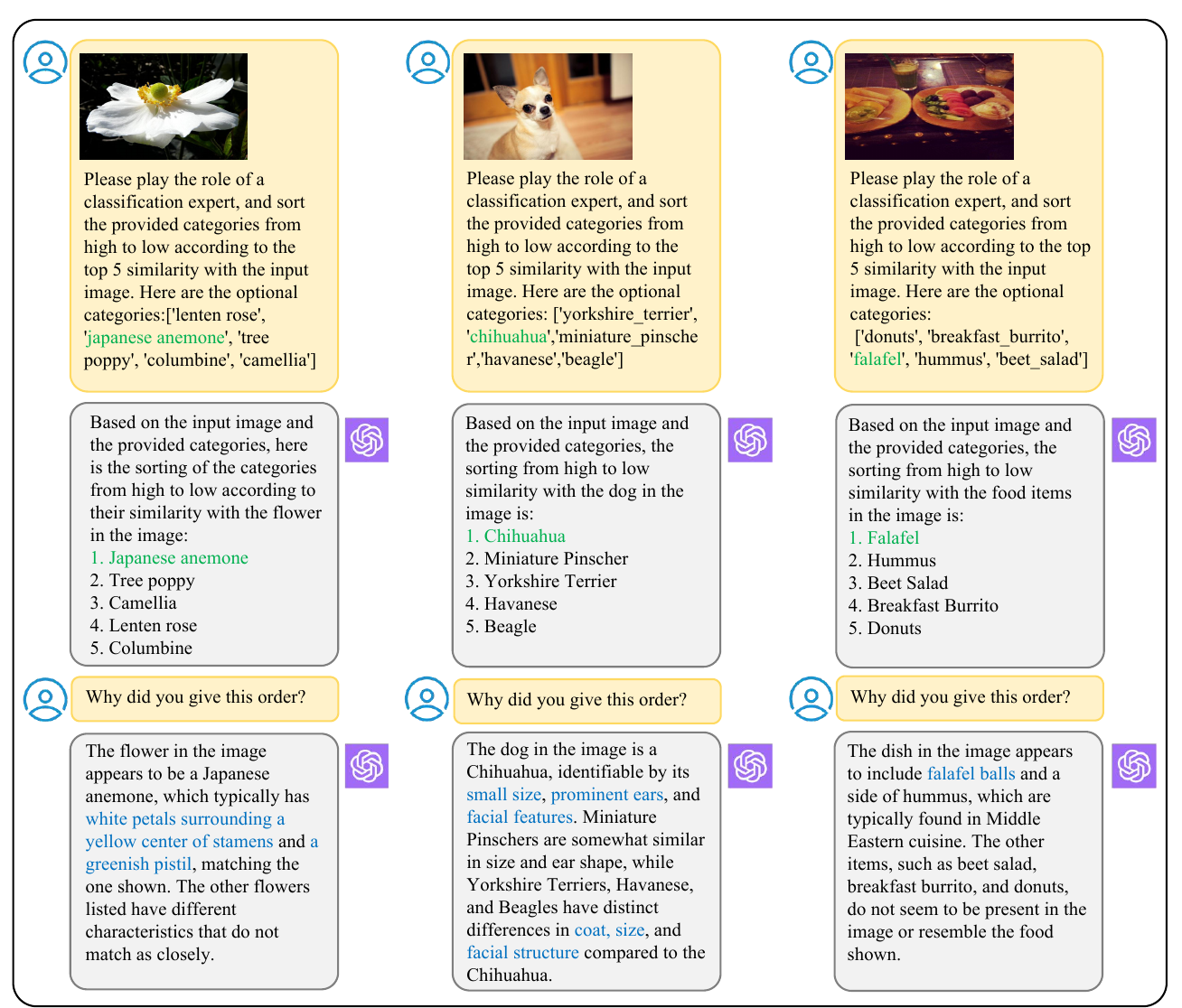}
  \vspace{-16pt}
  \caption{\textbf{GPT4V Example} for Flowers102, Pets37 and Food101. \hgreen{Green} for ground truth, \hblue{blue} for characteristics analyzed by GPT-4V. }
  \label{fig:4v2}
  \vspace{-16pt}
\end{figure}

\begin{figure}[t]
  \centering
  \includegraphics[width=1.\linewidth]{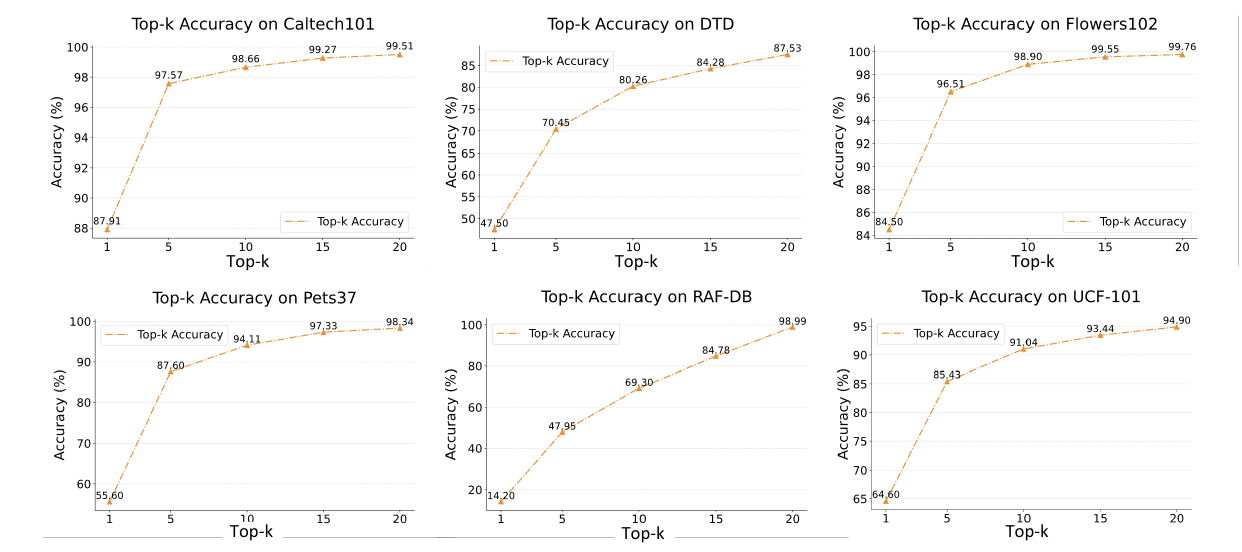}
  \vspace{-16pt}
  \caption{\textbf{Evaluation on CLIP+KNN} for Caltech101, Flowers102, RAF-DB, Pets37, DTD and UCF101. We report the top-1, 5, 10, 15, 20 accuracy (\%) under the 4-shot settings.}
  \label{fig:clip_fig}
  \vspace{-12pt}
\end{figure}

\begin{figure}[t]
  \centering
  \includegraphics[width=1.\linewidth]{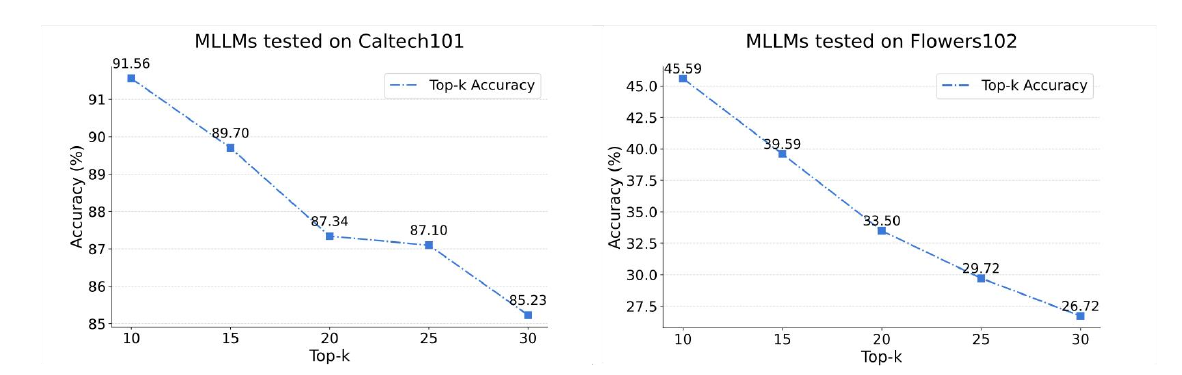}
  \vspace{-16pt}
  \caption{\textbf{Evaluation on MLLMs} for Caltech101, Flowers102. We report the test results using 10, 15, 20, 25, and 30 category names as inputs.}
  \label{fig:MLLMs}
  \vspace{-12pt}
\end{figure}

\noindent \textbf{More Discussion about Motivation.} 
To further demonstrate the potential of MLLMs in image classification tasks and to elaborate on the motivation behind our work, we employed the GPT-4V model to test selected images from our fine-grained datasets. Initially, we used the CLIP+KNN method to select 5 candidate images and their categories for a single image, ensuring that these candidates are at the top-5 in similarity among all images in memory, thus guaranteeing minimal differences between the chosen categories. Additionally, we intentionally selected examples that CLIP failed to classify correctly, increasing the complexity of the task. Subsequently, we presented these images and categories to GPT-4V, utilizing the prompt described in Sec.~\ref{sec:exp_setup}, prompting GPT-4V to rank all categories by similarity. During this process, we also requested GPT-4V to provide the rationale for its classifications, allowing us to analyze the specific role of MLLMs in classification tasks based on the reasons provided by GPT-4V. Fig.~\ref{fig:4v1} and Fig.~\ref{fig:4v2} presents several examples of five fine-grained classification datasets.

From the examples in Fig.~\ref{fig:4v1} and Fig.~\ref{fig:4v2}, it is evident that GPT-4V is capable of effectively analyzing the main feature information of objects in images during fine-grained image classification tasks. For instance, it identifies key characteristics such as “coupe” (a two-door car), “long fuselage” (long body of an aircraft), and “prominent ears” (noticeably protruding ears), which are crucial for distinguishing between similar categories. Sometimes, these detailed aspects may be overlooked by the CLIP model, leading to classification errors. Therefore, adopting a method of initial retrieval followed by deeper analysis, firstly filtering through the numerous fine-grained categories and then using MLLMs for further examination to select the most accurate answer, proves to be an effective approach for fine-grained image classification tasks.


Simultaneously, we assessed CLIP's accuracy in handling a variety of classification datasets. We selected six datasets: Caltech101, Flower102, RAF-DB, Pets37, DTD, and UCF101, and tested the CLIP+KNN method for top 1, 5, 10, 15, and 20 accuracy, with results presented in Fig.~\ref{fig:clip_fig}. We observed that as the top-k value increased, the classification accuracy improved rapidly, reaching over 90\% in four of the six datasets when top-k reached 10. This indicates that CLIP shows significant advantages as the number of predicted categories increases, complementing MLLMs' ability to discern among similar categories.

Following the experimental design in Fig.~\ref{fig:clip_fig}, we used MLLMs to rank categories when expanding the number of categories. We chose two datasets, Caltech101 and Flowers102, and used 10, 15, 20, 25, 30 categories as input to MLLMs, ensuring these included the correct category. As shown in Fig.~\ref{fig:MLLMs}, the distinction ability of MLLMs gradually decreased as the number of categories input into MLLMs increased.

Hence, we found that MLLMs and CLIP have complementary advantages in classification tasks. CLIP initially narrows down the correct answer to a smaller set through preliminary screening, while MLLMs can finely select the correct answer from this set. Our \methodname combines the strengths of both CLIP and MLLMs, first finding likely correct candidates through CLIP and retrieval, and then accurately selecting the correct answer through MLLMs' ranking, thus achieving outstanding results across multiple classification datasets.

From the 1-shot to 16-shot experiments in Tab.~\ref{tab:few_shot_appendix}, \methodname's results showed an improvement over the CLIP+KNN method by 7.4\%, 6.8\%, 6.2\%, 6.8\%, and 6.3\% respectively, averaging a 6.7\% percentage point increase, and significantly outperforming the performance of the LLaVa model itself. This outcome demonstrates the excellence of \methodname in image classification tasks (including fine-grained image classification), achieved by integrating the strengths of MLLMs and retrieval techniques.

\subsection{Zero-Shot Object Recognition}\label{sec:exp_zero_det}
Given the pre-existing object proposals such as ground-truth box annotations, the zero-shot object recognition task measures the model's capability of aligning regions with textual class descriptions.

\noindent \textbf{Baselines.}
We select two representative papers CLIP~\citep{radford2021learning} and RegionCLIP~\citep{zhong2022regionclip} and report their performances as the baseline results.
Besides, we apply our method on a range of cutting-edge open-source MLLMs, including LLaVA1.5~\citep{liu2023improved}, Qwen-VL~\citep{bai2023qwen} and InternLM-XC2~\citep{dong2024internlm}.

\noindent \textbf{Main Results on LVIS.} Tab.~\ref{tab:object_recognition} presents the results that reveal notable metrics improvements when applying our \methodname.
Specifically, when combing with the recent InternLM-XC2~\citep{dong2024internlm} model, our approach yielded an 8.4 (\%) point increase over the CLIP baseline and a 6.4 (\%) enhancement relative to RegionCLIP~\citep{zhong2022regionclip}.
These advancements underscore the efficacy of using an external memory for retrieval assistance coupled with the ranking prowess of MLLMs.

\noindent \textbf{Comparison with Rare Classes Results (\apr).} We find an interesting observation from the experimental results presented in Tab.~\ref{tab:object_recognition}. For the CLIP model, we observe a progressive increase in performance from \apr through \apc to \apf, which indicates a gradation in precision across varying class frequencies. However, employing our method yields a different trend, where \textit{the peak performance is achieved on \apr}, surpassing the CLIP model by as much as $19.6\%$. This significant leap in performance suggests a substantial advantage of our method when it comes to rare categories.
The integration of our \methodname to MLLMs plays a pivotal role here, as it demonstrates a heightened ability to discriminate among the rare classes.
Our observation could be attributed to the fact that our retrieving and reranking mechanism effectively pools relevant information from the external memory, providing the MLLMs with a richer context for rare class identification. Moreover, the ranking capability of MLLMs ensures that even the lesser-represented classes receive adequate attention during the classification process.
Our \methodname achieves a robust enhancement in the model's ability to discern and accurately classify objects that are infrequently encountered, addressing one of the significant challenges in long-tailed distribution datasets.

\begin{table}[t]
\small
\caption{
\footnotesize
Zero-shot object recognition on LVIS~\citep{gupta2019lvis} v1.0 \textit{validation} set.}
\label{tab:object_recognition}
  \centering
  \vspace{-6pt}
  \renewcommand{\arraystretch}{0.5}
  \resizebox{1.\linewidth}{!}{
  \begin{tabular}{@{}l|llll@{}}
  \toprule
   & \apr & \apc & \apf & \apall \\
  \cmidrule(r){1-1} \cmidrule(r){2-5}
  CLIP w/ box & 40.6 & 53.1 & \colorbox{secondBest}{59.2} & 48.7 \\
  CLIP w/ mask & 40.8 & 53.5 & \colorbox{firstBest}{\bf 59.6} & 49.2 \\
  RegionCLIP & 50.1 & 50.1 & 51.7 & 50.7  \\
  \cmidrule(r){1-1} \cmidrule(r){2-5}
  \methodname (LLaVA1.5) & 58.7 & \colorbox{secondBest}{57.9} & 54.4 & 56.2\\
  $\Delta$&\hgreen{+8.6}&\hgreen{+7.8}&\hgreen{+2.7}&\hgreen{+5.5}\\
  \cmidrule(r){1-1} \cmidrule(r){2-5}
  \methodname (Qwen-VL) & \colorbox{secondBest}{59.6} & 57.5 & 53.7 & \colorbox{secondBest}{56.4}\\
  $\Delta$&\hgreen{+9.5}&\hgreen{+7.4}&\hgreen{+2.0}&\hgreen{+5.7}\\
  \cmidrule(r){1-1} \cmidrule(r){2-5}
  \methodname (InternLM-XC2) & \colorbox{firstBest}{\bf 60.2} & \colorbox{firstBest}{\bf 58.0} & 54.3 & \colorbox{firstBest}{\bf 57.1} \\
  $\Delta$&\hgreen{+10.1}&\hgreen{+7.9}&\hgreen{+2.6}&\hgreen{+6.4}\\
  \bottomrule
  \end{tabular}}
  \vspace{-6pt}
\end{table}
\hspace{+1pt}
\begin{table}[t]
\small
\caption{
  \footnotesize Zero-shot object recognition on V3Det~\citep{wang2023v3det} \textit{validation} set with 13,204 categories.
  }
  \label{tab:object_recognition_v3det}
  \vspace{-6pt}
  \centering
  \renewcommand{\arraystretch}{0.5}
  \resizebox{1.\linewidth}{!}{
  \begin{tabular}{@{}l|llll@{}}
  \toprule
   & AP$_\textrm{s}$ & AP$_\textrm{m}$ & AP$_\textrm{l}$ & \apall \\
  \cmidrule(r){1-1} \cmidrule(r){2-5}
  CLIP w/ box & 7.2 & 12.9 & 12.8 & 9.8 \\
  \cmidrule(r){1-1} \cmidrule(r){2-5}
  \methodname (LLaVA1.5) & \colorbox{secondBest}{9.9} & \colorbox{firstBest}{\bf 13.2} & \colorbox{secondBest}{13.9} & \colorbox{secondBest}{11.1} \\
  $\Delta$&\hgreen{+2.7}&\hgreen{+0.3}&\hgreen{+1.1}&\hgreen{+1.3}\\
  \cmidrule(r){1-1} \cmidrule(r){2-5}
  \methodname (Qwen-VL) & 9.6 & 12.7 & 13.7 & 10.8 \\
  $\Delta$&\hgreen{+2.4}&\hblue{-0.2}&\hgreen{+0.9}&\hgreen{+1.0}\\
  \cmidrule(r){1-1} \cmidrule(r){2-5}
  \methodname (InternLM-XC2) & \colorbox{firstBest}{\bf 10.1} & \colorbox{secondBest}{13.1} & \colorbox{firstBest}{\bf 14.5} & \colorbox{firstBest}{\bf 11.3} \\
  $\Delta$&\hgreen{+2.9}&\hgreen{+0.2}&\hgreen{+1.7}&\hgreen{+1.5}\\
  \bottomrule
  \end{tabular}}
  \vspace{-6pt}
\end{table}

\noindent \textbf{Main Results on V3Det.} To further test the effectiveness of using MLLMs for ranking in scenarios with an extremely large number of fine-grained categories, we conducted additional experiments on V3Det~\citep{wang2023cogvlm}.
The experimental results in Tab.~\ref{tab:object_recognition_v3det} reveal that our \methodname has achieved a commendable improvement in performance, surpassing the CLIP baseline by 1.5 percentage points in overall average precision ($AP_{all}$) with InternLM-XC2. Such an improvement is particularly significant given the complexity of the V3Det dataset, which presents challenging 13,204 distinct classes. The MLLMs, with the aid of our retrieving and ranking mechanisms, have once again demonstrated their robust performance in the domain of object detection datasets. Using our retrieval-augmented approach allows MLLMs to navigate the extensive and fine-grained category landscape of V3Det effectively.

\noindent \textbf{Qualitative Results.}
\begin{figure*}[t]
  \centering
  \includegraphics[width=1.\linewidth]{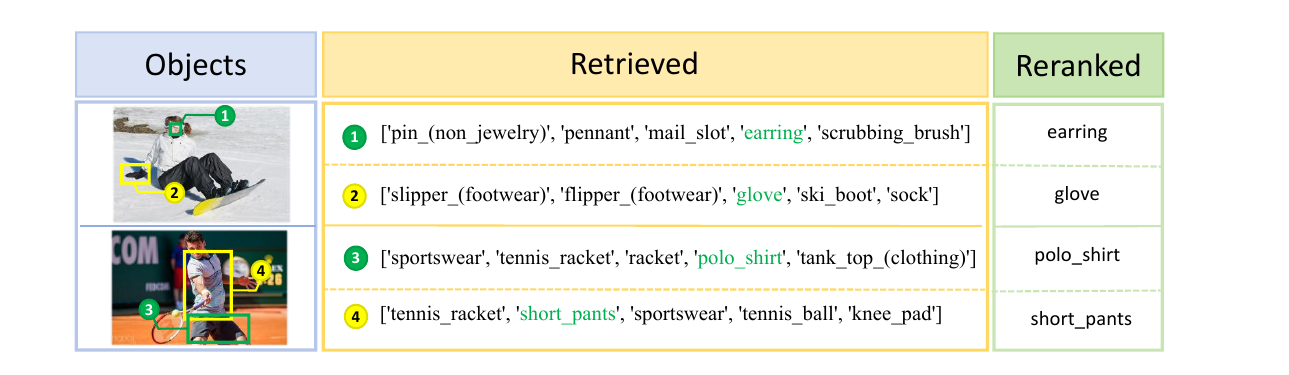}
  \vspace{-24pt}
  \caption{\textbf{Visualization of the ranking examples} for zero-shot object recognition on LVIS~\citep{gupta2019lvis} \textit{validation} set. Given the top retrieved predictions, our \methodname uses MLLMs to select the correct class names accurately.}
  \label{fig:rerank_example}
  \vspace{-12pt}
\end{figure*}
Fig.~\ref{fig:rerank_example} presents the visualization results about ranking examples of our approach on LVIS \textit{validation} set. The CLIP$\&$$K$-NN approach provides an extensive list of object predictions, albeit with the caveat that the most accurate label might not always emerge as the top-1 choice. The incorporation of MLLMs in our \methodname significantly streamlines the prediction process, yielding more precise and relevant object labels.
The visualization results demonstrate that our \methodname meets the need for fine-grained and large vocabulary recognition.

\subsection{Ablation Experiments}\label{sec:exp_ablation}

\begin{table*}[t]
  \setlength{\tabcolsep}{1mm}
  \footnotesize
  \centering
  \caption{
  \footnotesize
  \textbf{Ablation studies} about (1) using different datasets for fine-tuning and (2) fine-tuning $vs$ in-context learning. The symbols `F' and `S' stand for fine-tuning on the FGVC-Aircraft or Stanford-Cars datasets. 
  }
  \label{tab:ablation_datasets}
  \renewcommand{\arraystretch}{0.5}
  \resizebox{.95\linewidth}{!}{
  \begin{tabular}{l|ll| ccccccc ccc| c}
  \toprule
  Method & \multicolumn{2}{c|}{Strategy}  & \multicolumn{7}{c}{Common} 
  & \multicolumn{3}{c|}{Fine-Grained} \\
  \cmidrule(r){1-1} \cmidrule(r){2-3} \cmidrule(lr){4-10} \cmidrule(lr){11-13} \\
   & Fine-tune & In-Context & \rotbox{ImageNet} & \rotbox{Caltech101} & \rotbox{RAF-DB} & \rotbox{SUN397} & \rotbox{EuroSAT} & \rotbox{DTD} & \rotbox{UCF101} & \rotbox{Flower102} & \rotbox{Food101} & \rotbox{OxfordPets} & \rotbox{\textbf{Average}} \\
  \cmidrule(r){1-1} \cmidrule(r){2-3} \cmidrule(r){4-10} \cmidrule(r){11-13} \cmidrule(r){14-14}
  \multirow{2}{*}{\parbox{1.9cm}{\methodname}} & F & \xmark & \colorbox{firstBest}{\bf 75.8} & \colorbox{firstBest}{\bf 95.5} & \colorbox{firstBest}{\bf 66.0} & \colorbox{secondBest}{72.7} & \colorbox{firstBest}{\bf 90.7} & \colorbox{firstBest}{\bf 72.5} & \colorbox{firstBest}{\bf 81.4} & \colorbox{firstBest}{\bf 97.5} & \colorbox{secondBest}{88.1} & \colorbox{firstBest}{\bf 87.2} & \colorbox{firstBest}{\bf 82.7} \\
  ~ & S & \xmark & \colorbox{secondBest}{75.3} & \colorbox{secondBest}{94.9} & \colorbox{secondBest}{65.1} & \colorbox{firstBest}{\bf 73.1} & 88.1 & \colorbox{secondBest}{71.0} & \colorbox{secondBest}{81.1} & \colorbox{secondBest}{95.8} & \colorbox{firstBest}{\bf 88.3} & \colorbox{secondBest}{87.0} & \colorbox{secondBest}{82.0} \\
  (Qwen-VL) & \xmark & \cmark & 72.0 & 93.4 & 63.6 & 65.6 & 86.2 & 66.8 & 76.5 & 95.6 & 84.7 & 82.3 & 78.7 \\
  \cmidrule(r){1-1} \cmidrule(r){2-3} \cmidrule(r){4-10} \cmidrule(r){11-13} \cmidrule(r){14-14}
  \multirow{3}{*}{\parbox{1.9cm}{\methodname}} & F & \xmark & \colorbox{firstBest}{\bf 71.5} & \colorbox{secondBest}{94.4} & \colorbox{firstBest}{\bf 72.7} & \colorbox{firstBest}{\bf 69.7} & \colorbox{secondBest}{91.7} & \colorbox{firstBest}{\bf 69.9} & \colorbox{firstBest}{\bf 77.6} & \colorbox{secondBest}{93.2} & \colorbox{secondBest}{83.9} & \colorbox{secondBest}{79.3} & \colorbox{firstBest}{\bf 80.4} \\
  ~ & S & \xmark & \colorbox{firstBest}{\bf 71.5} & \colorbox{firstBest}{\bf 94.7} & \colorbox{secondBest}{71.2} & \colorbox{firstBest}{\bf 69.7} & 90.3 & \colorbox{firstBest}{\bf 69.9} & \colorbox{secondBest}{77.5} & 92.0 & 83.6 & \colorbox{firstBest}{\bf 79.7} & \colorbox{secondBest}{80.0} \\
  (InternLM-XC2) & \xmark & \cmark & 69.2 & 94.1 & 66.0 & \colorbox{firstBest}{\bf 69.7} & \colorbox{firstBest}{\bf 91.8} & 68.9 & 66.1 & \colorbox{firstBest}{\bf 95.7} & \colorbox{firstBest}{\bf 85.7} & 79.2 & 78.6 \\
  \bottomrule
  \end{tabular}
  }
\end{table*}

\begin{table}[t]
    \centering
    \footnotesize
    \caption{\textbf{Ablation studies} about the selection of the hyper-parameter $k$.}
    \renewcommand{\arraystretch}{0.5}
    \resizebox{1.\linewidth}{!}{
    \begin{tabular}{l|ccccc}
    \toprule
    & $k=3$ & $k=4$ & $k=5$ & $k=6$ & $k=7$ \\
    \cmidrule(r){1-1} \cmidrule(r){2-6}
     DTD & 70.27 & 71.34 & \colorbox{secondBest}{71.93} & \colorbox{secondBest}{71.93} & \colorbox{firstBest}{\bf 71.99} \\
     Flowers102 & \colorbox{firstBest}{\bf 96.18} & 95.57 & 95.62 & \colorbox{secondBest}{95.66} & 95.57 \\
     Oxford-pets & \colorbox{secondBest}{80.21} & \colorbox{firstBest}{\bf 80.38} & 79.91 & 79.72 & 79.42 \\
     Eurosat & 92.38 & 92.48 &  \colorbox{firstBest}{\bf 93.28} & 92.52 & \colorbox{secondBest}{92.59} \\
     \cmidrule(r){1-1} \cmidrule(r){2-6}
     Average & 84.76 & \colorbox{secondBest}{84.96} & \colorbox{firstBest}{\bf 85.19} & \colorbox{secondBest}{84.96} & 84.90 \\
     \bottomrule
    \end{tabular}}
    \hspace{+2pt}
    \footnotesize
    \label{tab:exps_knn}
    \vspace{-24pt}
\end{table}

\begin{figure}[t]
  \centering
  \includegraphics[width=.95\linewidth]{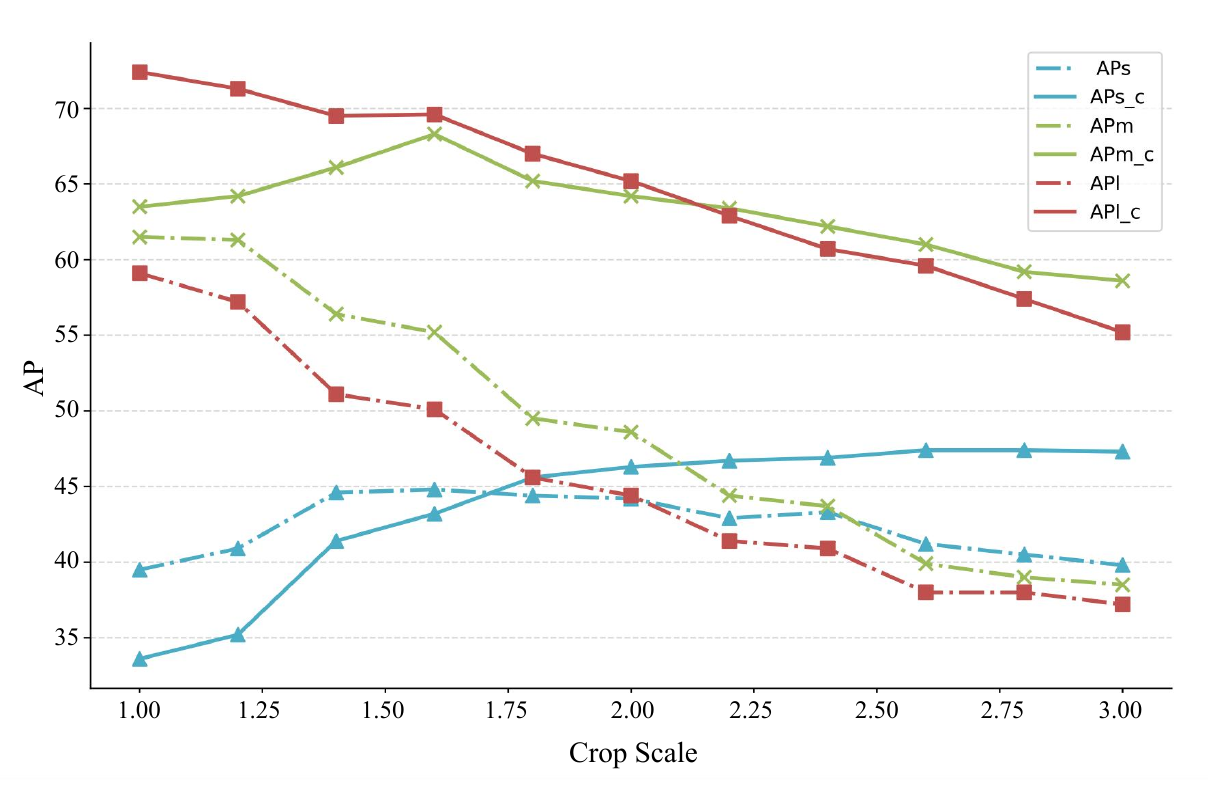}
  \vspace{-16pt}
  \caption{\textbf{Metric curve visualization of CLIP~\citep{radford2021learning} zero-shot classification} on LVIS~\citep{gupta2019lvis} with ground truth proposals. Different behaviors can be seen before and after blurring with respect to different object's scales.}
  \label{fig:lvis_ablation}
  \vspace{-16pt}
\end{figure}

\begin{table*}[ht]
\setlength{\tabcolsep}{1mm}
\footnotesize
\caption{Reuslts on more baselines. The experimental results are based on the 4-shot setting.}
\label{tab:More}
\renewcommand{\arraystretch}{1.2}
\centering
\resizebox{\textwidth}{!}{
\begin{tabular}{l|ccccccc|cccc|c}
\toprule
\textbf{Model} & \multicolumn{7}{c|}{\textbf{Common}} & \multicolumn{4}{c|}{\textbf{Fine-Grained}} & \textbf{Average} \\
\cmidrule(r){2-8} \cmidrule(r){9-12} \cmidrule(r){13-13}
 & \textbf{ImageNet} & \textbf{Caltech101} & \textbf{RAF-DB} & \textbf{SUN397} & \textbf{EuroSAT} & \textbf{DTD} & \textbf{UCF-101} 
 & \textbf{Flower102} & \textbf{StanfordCars} & \textbf{Food101} & \textbf{OxfordPets} & \textbf{} \\
\midrule
EVA-CLIP (ViT-L/14) & 58.2 & 93.6 & 27.8 & 58.8 & 78.8 & 56.3 & 76.3 & 97.6 & 79.1 & 78.1 & 73.9 & 70.8 \\
\rowcolor{GrayBG} {\methodname(Qwen2.5-VL)} & 72.1 & 96.0 & 43.5 & 72.4 & 71.6 & 67.0 & 81.8 & 92.3 & 87.0 & 85.9 & 89.5 & 78.1 \\
$\Delta$ & \hgreen{+13.9} & \hgreen{+2.4} & \hgreen{+15.7} & \hgreen{+13.6} & \hblue{-7.2} & \hgreen{+10.7} & \hgreen{+5.5} & \hblue{-5.3} & \hgreen{+7.9} & \hgreen{+7.8} & \hgreen{+15.6} & \hgreen{+7.3}\\
\hline
OpenCLIP (ViT-g/14) & 54.5 & 93.6 & 29.1 & 59.2 & 75.5 & 58.6 & 75.8 & 93.8 & 78.8 & 72.4 & 74.4 & 69.6 \\
\rowcolor{GrayBG} \methodname(Qwen2.5-VL) & 70.2 & 95.5 & 45.0 & 72.2 & 73.5 & 69.4 & 82.2 & 91.2 & 87.9 & 83.8 & 88.7 & 78.1 \\
$\Delta$ & \hgreen{+15.7} & \hgreen{+1.9} & \hgreen{+15.9} & \hgreen{+13.0} & \hblue{-2.0} & \hgreen{+10.8} & \hgreen{+6.4} & \hblue{-2.6} & \hgreen{+9.1} & \hgreen{+11.4} & \hgreen{+14.3} & \hgreen{+8.5} \\
\hline
OpenCLIP (ViT-B/16) & 39.8 & 86.2 & 14.9 & 48.4 & 63.8 & 41.3 & 59.8 & 80.6 & 44.3 & 58.6 & 54.8 & 53.9 \\
\rowcolor{GrayBG} \methodname(Qwen2.5-VL) & 61.2 & 93.4 & 33.2 & 66.3 & 57.5 & 57.2 & 75.4 & 86.3 & 71.5 & 77.8 & 82.2 & 69.3 \\
$\Delta$ & \hgreen{+21.4} & \hgreen{+7.2} & \hgreen{+18.3} & \hgreen{+17.9} & \hblue{-6.3} & \hgreen{+15.9} & \hgreen{+15.6} & \hgreen{+5.7} & \hgreen{+27.2} & \hgreen{+19.2} & \hgreen{+27.4} & \hgreen{+15.4} \\
\hline
OpenCLIP (RN50) & 31.0 & 80.2 & 22.2 & 43.7 & 60.6 & 43.7 & 55.7 & 72.0 & 35.4 & 58.6 & 54.8 & 48.1 \\
\rowcolor{GrayBG} \methodname(Qwen2.5-VL) & 54.9 & 92.6 & 36.8 & 63.7 & 59.6 & 59.6 & 73.0 & 84.9 & 71.5 & 77.8 & 82.2 & 69.3 \\
$\Delta$ & \hgreen{+23.9} & \hgreen{+12.4} & \hgreen{+14.6} & \hgreen{+20.0} & \hblue{-1.0} & \hgreen{+15.9} & \hgreen{+17.3} & \hgreen{+12.9} & \hgreen{+36.1} & \hgreen{+19.2} & \hgreen{+27.4} & \hgreen{+21.2} \\
\bottomrule
\end{tabular}
}
\end{table*}

\begin{table}[ht]
\centering
\caption{Ablation study on HNSW and brute-force retrieval methods on three datasets. The experimental results are based on the 4-shot setting. We traversed the entire test set and calculated the average time for a single retrieval.}
\resizebox{1.\linewidth}{!}{
\begin{tabular}{l|c|c|c|c|c|c}
\toprule
\multirow{2}{*}{\textbf{Model}} & \multicolumn{2}{c|}{\textbf{Caltech101}} & \multicolumn{2}{c|}{\textbf{Oxford-Flowers}} & \multicolumn{2}{c}{\textbf{DTD}} \\
\cline{2-7}
 & T(ms) & Acc & T(ms) & Acc & T(ms) & Acc \\
\hline
ViT-B/16+Brute-force & 2.2 & 90.4 & 2.1 & 70.5 & 1.3 & 43.2 \\
\rowcolor{GrayBG} ViT-B/16+HNSW & 0.3 & 86.2 & 0.3 & 80.6 & 0.2 & 41.3 \\
$\Delta$ & $+\downarrow86.4\%$ & -4.2 & $+\downarrow85.7\%$ & +10.1 & $+\downarrow84.6\%$ & -1.9 \\
\bottomrule
\end{tabular}
}
\label{tab:brute force ablation}
\end{table}

\begin{table}[ht]
\centering
\caption{Ablation study on different retrieval methods. The experimental results are based on the 4-shot setting.}
\begin{tabular}{lcccc}
\toprule
\textbf{Dataset} & \textbf{IndexHNSW} & \textbf{IndexPQ} & \textbf{IndexIVF} & \textbf{IndexLSH} \\
\midrule
Caltech101 & 86.2 & 86.2 & 81.7 & 29.2 \\
DTD & 41.3 & 30.1 & 35.3 & 11.1 \\
\bottomrule
\end{tabular}
\label{tab:ablation on retrieval methods}
\end{table}

\begin{table}[h]
\small
\caption{
Zero-shot object recognition on the LVIS v1.0 validation set, comparing performance between ground-truth proposals and Detic-generated proposals.}
\label{tab:detic_proposal}
  \centering
  \renewcommand{\arraystretch}{0.5}
  \resizebox{.95\linewidth}{!}{
  \begin{tabular}{@{}l|l|llll@{}}
  \toprule
   & Proposal & $AP_{r}$ & $AP_{c}$ & $AP_{f}$ & $AP_{all}$ \\
  \cmidrule(r){1-1} \cmidrule(r){2-2} \cmidrule(r){3-6}
  CLIP w/ box & GT & 40.6 & 53.1 & 59.2 & 48.7 \\
  \methodname & GT & 58.7 & 57.9 & 54.4 & 56.2\\
  \cmidrule(r){1-1} \cmidrule(r){2-2} \cmidrule(r){3-6}
  CLIP w/ box & Detic & 13.3 & 13.5 & 13.3 & 13.4 \\
  \methodname & Detic & 25.7 & 24.1 & 11.7 & 19.2\\
  \bottomrule
  \end{tabular}
}
\end{table}

\noindent \textbf{Different Fine-tuning data.} We study the importance of using different fine-tuning datasets for ranking.
We select two representative datasets: FGVC-Aircraft and Stanford-Cars as the data sources for constructing the fine-tuning data.
Our selection is motivated by their diverse characteristics and relevance in visual recognition tasks, providing a comprehensive basis for fine-tuning.
Subsequently, we fine-tune the \methodname with different MLLMs (Qwen-VL and InternLM-XC2) on these two datasets, aiming to investigate how different data sources influence performance.
To thoroughly assess the impact of using different fine-tuning datasets, we evaluate the fine-tuned \methodname across a diverse set of 10 additional datasets.

Tab.~\ref{tab:ablation_datasets} presents the results. We observe that \methodname is not sensitive to changes in the fine-tuning dataset for ranking, thereby confirming its viability as a generalizable and reliable method for enhancing the performance of MLLMs.
The consistency in results, irrespective of the fine-tuning data source, underlines the robustness of our fine-tuning strategy.
Despite these minor variations, the overall performance of using FGVC-Aircrafts (82.7$\%$, top row) is higher than using StanfordCars (82.0$\%$, second row) for Qwen-VL, and we observe the same trend for InternLM-XC2.
Based on our findings, we adopt the FGVC-Aircraft dataset as our preferred choice for fine-tuning.

\noindent \textbf{Fine-tuning $vs$ In-Context Learning.}
We validate the effectiveness of fine-tuning the MLLM or just in-context learning (training-free) for ranking. The results are illustrated in Tab.~\ref{tab:ablation_datasets}. We select two distinct groups for comparison. The first group (top and fourth rows) involves models that are fine-tuned using the FGVC-Aircraft dataset, while the second group (third and bottom rows) consists of models with in-context learning prompts for ranking. The results show a consistent improvement in accuracy for the fine-tuned model across almost all datasets for both Qwen-VL and InternLM-XC2. The notable enhancement in performance across a diverse range of datasets highlights the efficacy of our fine-tuning strategy. The results substantiate that fine-tuning the MLLM with target datasets like FGVC-Aircraft significantly bolsters the model's ranking capabilities.

\noindent \textbf{Ablation Study on Effects of the Parameter $k$.} We delve into the impact of the hyper-parameter $k$ on few-shot image recognition setting, as detailed in Tab.~\ref{tab:exps_knn}.
We report the results of \methodname with the LLaVA1.5 as the MLLM.
The average results obtained from different values of $k$ varied from a minimum of 84.8 to a maximum of 85.2, with a fluctuation range of only 0.4 points. Compared to the model's performance improvement, this fluctuation range is relatively small. Our findings reveal that our \methodname demonstrates a remarkable robustness to variations in $k$, with only minor differences observed across a broad spectrum of values from $3$ to $7$.
Such a consistency suggests that \methodname's ability to generalize from a few examples is not significantly influenced by the choice of $k$.
We do not rely on carefully selected parameters on the test set to achieve significant performance improvements. For a reasonable range of $k$, RAR consistently delivers substantial performance gains.

\noindent \textbf{Results on More MLLM and Embedding Models.} We further evaluated RAR by replacing the retriever with both stronger and weaker embedding models. On the one hand, we employed more powerful embedding models, such as EVA-CLIP~\citep{EVA-CLIP} and OpenCLIP~\citep{ilharco_gabriel_2021_5143773} with a ViT-g/14 backbone, to verify the effectiveness of RAR under higher baselines. On the other hand, we also tested weaker embedding models, such as CLIP with a ResNet50~\citep{He2015DeepRL} backbone, to examine the impact of reduced retriever capability. Compared with the MLLMs used in Tab.\ref{tab:few_shot_appendix} and Tab.\ref{tab:4v_classification} (e.g., LLaVA and Qwen-VL), we additionally incorporated the more advanced Qwen2.5-VL-7B model~\citep{bai2025qwen2}. To ensure fairness, Qwen2.5-VL-7B was fine-tuned with exactly the same training parameters and datasets.

All experiments were conducted under the 4-shot setting, and the results are summarized in Tab.~\ref{tab:More}. As shown in the table, employing stronger embedding models naturally increases the baseline: from OpenCLIP~\citep{ilharco_gabriel_2021_5143773} (ViT-B/16, 53.9) to OpenCLIP~\citep{ilharco_gabriel_2021_5143773} (ViT-g/14, 69.6) and further to EVA-CLIP~\citep{EVA-CLIP} (ViT-L/14, 70.8). Importantly, RAR consistently delivers significant improvements on top of these stronger baselines, with gains of +15.4, +8.5, and +7.3, respectively. This demonstrates that RAR remains effective even when the baseline has already reached a strong level (e.g., EVA-CLIP). Conversely, with the weaker OpenCLIP (RN50) embedding model, which produces a lower baseline (48.1), RAR achieves a much larger improvement of +21.2 on average.

These results highlight two important observations: (1) as the capability of the embedding model increases or decreases, the baseline performance shifts accordingly; and (2) RAR consistently yields significant gains regardless of the underlying retriever strength. The inclusion of embedding models with different capacity levels, together with a stronger MLLM, further enriches our experimental baselines and validates the robustness and generalization ability of RAR across diverse settings.

\noindent \textbf{Ablation study on HNSW and brute-force retrieval methods on three datasets.}
To clarify the comparison between HNSW and brute-force retrieval in the context of \methodname, we analyzed their computational characteristics and conducted experiments. In summary, brute-force retrieval scales linearly with the dataset size, while HNSW achieves logarithmic search complexity by leveraging a hierarchical graph structure. As a result, HNSW offers substantially faster retrieval efficiency across datasets.

Additionally, we compared retrieval methods based on HNSW and brute-force search, and calculated the average time for a single retrieval. Table~\ref{tab:brute force ablation} presents the results of our tests on several datasets. Our ablation experiments were conducted using the embedding model OpenCLIP (ViT-B/16), and we evaluated the CLIP+KNN experimental results. Our findings are: 1) the HNSW-based retrieval method is significantly faster than brute-force search; 2) using HNSW results in a drop in retrieval accuracy on some datasets (e.g., Caltech101), while it improves retrieval accuracy on others (e.g., Oxford-Flowers).

These experiments validate the significant advantage of HNSW in accelerating the retrieval process, while also showing the potential trade-offs in accuracy on certain datasets.

\noindent \textbf{Ablation study on different retrieval methods.} To further investigate the influence of the retriever’s capability on overall model performance, we have added several alternative indexing and retrieval methods supported by FAISS~\citep{douze2024faiss}, such as IndexHNSW, IndexPQ, IndexIVF, and IndexLSH. We then tested the top-1 retrieval accuracy on two datasets using these methods. The results are in Tab.~\ref{tab:ablation on retrieval methods}. As shown in Tab.~\ref{tab:ablation on retrieval methods}, the method used by \methodname already achieves near-optimal retrieval accuracy. Other methods, such as IndexIVF and IndexLSH, yield lower results, especially on the DTD dataset, where IndexLSH performs significantly worse. These findings suggest that the current retrieval method is already highly effective, and additional other retrieval methods have not provided significant improvements in our experiments.

\noindent \textbf{Ablation study on different proposals.} In our experiments (Tab.~\ref{tab:object_recognition} and Tab.~\ref{tab:object_recognition_v3det}), all models use the same ground-truth bounding boxes as input. Among them, \textit{CLIP w/mask} is the only method that uses ground-truth masks from the LVIS dataset.
To further clarify, we have added an additional experiment using proposals generated by Detic~\citep{Detic}. As show in Tab.~\ref{tab:detic_proposal}, the overall performance trends remain consistent, our RAR still demonstrates a clear improvement over baseline. However, since Detic~\citep{Detic} proposals contain some inaccurate detections, the absolute AP scores are lower than those obtained with ground-truth bounding boxes.

\section{Conclusion}
\vspace{-6pt}
In this paper, we highlight the potential of combining retrieving and ranking with multi-modal large language models to revolutionize perception tasks such as fine-grained recognition, zero-shot image recognition, and few-shot object recognition.
Motivated by the limited zero-shot/few-shot of CLIP and MLLMs on fine-grained datasets, our \methodname designs the pipeline that uses MLLM to rank the retrieved results.
Our proposed approach can be seamlessly integrated into various MLLMs for real-world applications where the variety and volume of categories continuously expand.
Our method opens up new avenues for research in augmenting the MLLM's abilities with the retrieving-augmented solution and could be beneficial for other tasks such as reasoning and generation in future works.

\section*{Acknowledgments}
This work was supported by National Key R$\&$D Program of China 2022ZD0161600, Shanghai Artificial lntelligence Laboratory, Hong Kong RGC TRS T41-603/20-R, the Centre for Perceptual and Interactive Intelligence (CPII) Ltd under the Innovation and Technology Commission (ITC)’s InnoHK. Dahua Lin is a PI of CPII under the InnoHK.

\bibliographystyle{IEEEtran}  
\bibliography{ieee}  

@inproceedings{gupta2019lvis,
  title={{LVIS}: A dataset for large vocabulary instance segmentation},
  author={Gupta, Agrim and Dollar, Piotr and Girshick, Ross},
  booktitle={CVPR},
  year={2019}
}

@inproceedings{radford2021learning,
  title={Learning transferable visual models from natural language supervision},
  author={Radford, Alec and Kim, Jong Wook and Hallacy, Chris and Ramesh, Aditya and Goh, Gabriel and Agarwal, Sandhini and Sastry, Girish and Askell, Amanda and Mishkin, Pamela and Clark, Jack and others},
  booktitle={ICML},
  pages={8748--8763},
  year={2021},
}

@article{wu2023gpt4vis,
  title={{GPT4Vis}: What Can GPT-4 Do for Zero-shot Visual Recognition?},
  author={Wu, Wenhao and Yao, Huanjin and Zhang, Mengxi and Song, Yuxin and Ouyang, Wanli and Wang, Jingdong},
  journal={arXiv preprint arXiv:2311.15732},
  year={2023}
}

@misc{2023gpt4vision,
  title={{GPT-4V(ision)} System Card},
  author={OpenAI},
  year={2023},
  url={https://openai.com/research/gpt-4v-system-card}
}

@article{zhang2023internlm,
  title={{Internlm-Xcomposer}: A vision-language large model for advanced text-image comprehension and composition},
  author={Zhang, Pan and Wang, Xiaoyi Dong Bin and Cao, Yuhang and Xu, Chao and Ouyang, Linke and Zhao, Zhiyuan and Ding, Shuangrui and Zhang, Songyang and Duan, Haodong and Yan, Hang and others},
  journal={arXiv preprint arXiv:2309.15112},
  year={2023}
}

@article{liu2023improved,
  title={Improved baselines with visual instruction tuning},
  author={Liu, Haotian and Li, Chunyuan and Li, Yuheng and Lee, Yong Jae},
  journal={arXiv preprint arXiv:2310.03744},
  year={2023}
}

@inproceedings{cimpoi2014describing,
  title={Describing textures in the wild},
  author={Cimpoi, Mircea and Maji, Subhransu and Kokkinos, Iasonas and Mohamed, Sammy and Vedaldi, Andrea},
  booktitle={CVPR},
  year={2014}
}

@article{helber2019eurosat,
  title={Eurosat: A novel dataset and deep learning benchmark for land use and land cover classification},
  author={Helber, Patrick and Bischke, Benjamin and Dengel, Andreas and Borth, Damian},
  journal={IEEE J. Sel. Top. Appl. Earth Obs. Remote Sens.},
  year={2019},
}

@inproceedings{fei2004learning,
  title={Learning generative visual models from few training examples: An incremental bayesian approach tested on 101 object categories},
  author={Fei-Fei, Li and Fergus, Rob and Perona, Pietro},
  booktitle={CVPR workshop},
  year={2004},
}

@article{maji2013fine,
  title={Fine-grained visual classification of aircraft},
  author={Maji, Subhransu and Rahtu, Esa and Kannala, Juho and Blaschko, Matthew and Vedaldi, Andrea},
  journal={arXiv preprint arXiv:1306.5151},
  year={2013}
}

@inproceedings{xiao2010sun,
  title={{SUN} database: Large-scale scene recognition from abbey to zoo},
  author={Xiao, Jianxiong and Hays, James and Ehinger, Krista A and Oliva, Aude and Torralba, Antonio},
  booktitle={CVPR},
  year={2010},
}

@inproceedings{nilsback2008automated,
  title={Automated flower classification over a large number of classes},
  author={Nilsback, Maria-Elena and Zisserman, Andrew},
  booktitle={ICVGIP},
  year={2008},
}

@inproceedings{parkhi2012cats,
  title={Cats and dogs},
  author={Parkhi, Omkar M and Vedaldi, Andrea and Zisserman, Andrew and Jawahar, CV},
  booktitle={CVPR},
  year={2012},
}

@inproceedings{krause20133d,
  title={3d object representations for fine-grained categorization},
  author={Krause, Jonathan and Stark, Michael and Deng, Jia and Fei-Fei, Li},
  booktitle={ICCV workshops},
  year={2013}
}

@inproceedings{li2017reliable,
  title={Reliable crowdsourcing and deep locality-preserving learning for expression recognition in the wild},
  author={Li, Shan and Deng, Weihong and Du, JunPing},
  booktitle={CVPR},
  year={2017}
}

@inproceedings{jia2021scaling,
  title={Scaling up visual and vision-language representation learning with noisy text supervision},
  author={Jia, Chao and Yang, Yinfei and Xia, Ye and Chen, Yi-Ting and Parekh, Zarana and Pham, Hieu and Le, Quoc and Sung, Yun-Hsuan and Li, Zhen and Duerig, Tom},
  booktitle={ICML},
  year={2021},
}

@inproceedings{fang2023eva,
  title={{EVA}: Exploring the limits of masked visual representation learning at scale},
  author={Fang, Yuxin and Wang, Wen and Xie, Binhui and Sun, Quan and Wu, Ledell and Wang, Xinggang and Huang, Tiejun and Wang, Xinlong and Cao, Yue},
  booktitle={CVPR},
  year={2023}
}

@article{sun2023alpha,
  title={{Alpha-CLIP}: A CLIP Model Focusing on Wherever You Want},
  author={Sun, Zeyi and Fang, Ye and Wu, Tong and Zhang, Pan and Zang, Yuhang and Kong, Shu and Xiong, Yuanjun and Lin, Dahua and Wang, Jiaqi},
  journal={arXiv preprint arXiv:2312.03818},
  year={2023}
}

@inproceedings{li2022blip,
  title={{BLIP}: Bootstrapping language-image pre-training for unified vision-language understanding and generation},
  author={Li, Junnan and Li, Dongxu and Xiong, Caiming and Hoi, Steven},
  booktitle={ICML},
  year={2022},
}

@inproceedings{li2023blip,
  title={{BLIP-2}: Bootstrapping language-image pre-training with frozen image encoders and large language models},
  author={Li, Junnan and Li, Dongxu and Savarese, Silvio and Hoi, Steven},
  booktitle={ICML},
  year={2023}
}

@inproceedings{li2022grounded,
  title={Grounded language-image pre-training},
  author={Li, Liunian Harold and Zhang, Pengchuan and Zhang, Haotian and Yang, Jianwei and Li, Chunyuan and Zhong, Yiwu and Wang, Lijuan and Yuan, Lu and Zhang, Lei and Hwang, Jenq-Neng and others},
  booktitle={CVPR},
  year={2022}
}

@inproceedings{zhong2022regionclip,
  title={{RegionCLIP}: Region-based language-image pretraining},
  author={Zhong, Yiwu and Yang, Jianwei and Zhang, Pengchuan and Li, Chunyuan and Codella, Noel and Li, Liunian Harold and Zhou, Luowei and Dai, Xiyang and Yuan, Lu and Li, Yin and others},
  booktitle={CVPR},
  year={2022}
}

@inproceedings{gu2021open,
  title={Open-vocabulary object detection via vision and language knowledge distillation},
  author={Gu, Xiuye and Lin, Tsung-Yi and Kuo, Weicheng and Cui, Yin},
  booktitle={ICLR},
  year={2022}
}

@inproceedings{zhou2022detecting,
  title={Detecting twenty-thousand classes using image-level supervision},
  author={Zhou, Xingyi and Girdhar, Rohit and Joulin, Armand and Kr{\"a}henb{\"u}hl, Philipp and Misra, Ishan},
  booktitle={ECCV},
  year={2022},
}

@article{gao2023clip,
  title={{Clip-Adapter}: Better vision-language models with feature adapters},
  author={Gao, Peng and Geng, Shijie and Zhang, Renrui and Ma, Teli and Fang, Rongyao and Zhang, Yongfeng and Li, Hongsheng and Qiao, Yu},
  journal={IJCV},
  year={2023},
}

@article{zhou2021coop,
    title={Learning to Prompt for Vision-Language Models},
    author={Zhou, Kaiyang and Yang, Jingkang and Loy, Chen Change and Liu, Ziwei},
    journal={IJCV},
    year={2022}
}

@article{zhu2023minigpt,
  title={{MiniGPT-4}: Enhancing vision-language understanding with advanced large language models},
  author={Zhu, Deyao and Chen, Jun and Shen, Xiaoqian and Li, Xiang and Elhoseiny, Mohamed},
  journal={arXiv preprint arXiv:2304.10592},
  year={2023}
}

@inproceedings{liu2024visual,
  title={Visual instruction tuning},
  author={Liu, Haotian and Li, Chunyuan and Wu, Qingyang and Lee, Yong Jae},
  booktitle={NeurIPS},
  year={2024}
}

@misc{dai2023instructblip,
      title={InstructBLIP: Towards General-purpose Vision-Language Models with Instruction Tuning}, 
      author={Wenliang Dai and Junnan Li and Dongxu Li and Anthony Meng Huat Tiong and Junqi Zhao and Weisheng Wang and Boyang Li and Pascale Fung and Steven Hoi},
      year={2023},
      eprint={2305.06500},
      journal={arXiv},
      primaryClass={cs.CV}
}

@article{ye2023mplug,
  title={mplug-owl: Modularization empowers large language models with multimodality},
  author={Ye, Qinghao and Xu, Haiyang and Xu, Guohai and Ye, Jiabo and Yan, Ming and Zhou, Yiyang and Wang, Junyang and Hu, Anwen and Shi, Pengcheng and Shi, Yaya and others},
  journal={arXiv},
  year={2023}
}

@article{bai2023qwen,
  title={{Qwen-VL}: A Frontier Large Vision-Language Model with Versatile Abilities},
  author={Bai, Jinze and Bai, Shuai and Yang, Shusheng and Wang, Shijie and Tan, Sinan and Wang, Peng and Lin, Junyang and Zhou, Chang and Zhou, Jingren},
  journal={arXiv},
  year={2023}
}

@article{peng2023kosmos,
  title={Kosmos-2: Grounding Multimodal Large Language Models to the World},
  author={Peng, Zhiliang and Wang, Wenhui and Dong, Li and Hao, Yaru and Huang, Shaohan and Ma, Shuming and Wei, Furu},
  journal={arXiv},
  year={2023}
}

@article{awadalla2023openflamingo,
  title={OpenFlamingo: An Open-Source Framework for Training Large Autoregressive Vision-Language Models},
  author={Anas Awadalla and Irena Gao and Josh Gardner and Jack Hessel and Yusuf Hanafy and Wanrong Zhu and Kalyani Marathe and Yonatan Bitton and Samir Gadre and Shiori Sagawa and Jenia Jitsev and Simon Kornblith and Pang Wei Koh and Gabriel Ilharco and Mitchell Wortsman and Ludwig Schmidt},
  journal={arXiv},
  year={2023}
}

@misc{wang2023cogvlm,
      title={CogVLM: Visual Expert for Pretrained Language Models}, 
      author={Weihan Wang and Qingsong Lv and Wenmeng Yu and Wenyi Hong and Ji Qi and Yan Wang and Junhui Ji and Zhuoyi Yang and Lei Zhao and Xixuan Song and Jiazheng Xu and Bin Xu and Juanzi Li and Yuxiao Dong and Ming Ding and Jie Tang},
      year={2023},
      eprint={2311.03079},
      archivePrefix={arXiv},
      primaryClass={cs.CV}
}

@article{chen2023sharegpt4v,
  title={Sharegpt4v: Improving large multi-modal models with better captions},
  author={Chen, Lin and Li, Jisong and Dong, Xiaoyi and Zhang, Pan and He, Conghui and Wang, Jiaqi and Zhao, Feng and Lin, Dahua},
  journal={arXiv preprint arXiv:2311.12793},
  year={2023}
}

@inproceedings{yasunaga2022retrieval,
  title={Retrieval-augmented multimodal language modeling},
  author={Yasunaga, Michihiro and Aghajanyan, Armen and Shi, Weijia and James, Rich and Leskovec, Jure and Liang, Percy and Lewis, Mike and Zettlemoyer, Luke and Yih, Wen-tau},
  booktitle={ICML},
  year={2023}
}

@inproceedings{deng2009imagenet,
  title={{ImageNet}: A large-scale hierarchical image database},
  author={Deng, Jia and Dong, Wei and Socher, Richard and Li, Li-Jia and Li, Kai and Fei-Fei, Li},
  booktitle={CVPR},
  year={2009},
}

@article{yang2023recognize,
  title={Recognize any regions},
  author={Yang, Haosen and Ma, Chuofan and Wen, Bin and Jiang, Yi and Yuan, Zehuan and Zhu, Xiatian},
  journal={arXiv preprint arXiv:2311.01373},
  year={2023}
}

@inproceedings{wang2023v3det,
  author       = {Jiaqi Wang and
                  Pan Zhang and
                  Tao Chu and
                  Yuhang Cao and
                  Yujie Zhou and
                  Tong Wu and
                  Bin Wang and
                  Conghui He and
                  Dahua Lin},
  title        = {{V3Det}: Vast Vocabulary Visual Detection Dataset},
  booktitle    = {ICCV},
  year         = {2023},
}

@article{malkov2018efficient,
  title={Efficient and robust approximate nearest neighbor search using hierarchical navigable small world graphs},
  author={Malkov, Yu A and Yashunin, Dmitry A},
  journal={TPAMI},
  year={2018},
}

@article{dong2024internlm,
  title={{InternLM-XComposer2}: Mastering free-form text-image composition and comprehension in vision-language large model},
  author={Dong, Xiaoyi and Zhang, Pan and Zang, Yuhang and Cao, Yuhang and Wang, Bin and Ouyang, Linke and Wei, Xilin and Zhang, Songyang and Duan, Haodong and Cao, Maosong and others},
  journal={arXiv preprint arXiv:2401.16420},
  year={2024}
}

@inproceedings{conti2024vocabulary,
  title={Vocabulary-free image classification},
  author={Conti, Alessandro and Fini, Enrico and Mancini, Massimiliano and Rota, Paolo and Wang, Yiming and Ricci, Elisa},
  booktitle={NeurIPS},
  year={2024}
}

@inproceedings{liu2024democratizing,
  title={Democratizing Fine-grained Visual Recognition with Large Language Models},
  author={Liu, Mingxuan and Roy, Subhankar and Li, Wenjing and Zhong, Zhun and Sebe, Nicu and Ricci, Elisa},
  booktitle={ICLR},
  year={2024}
}

@article{miller1995wordnet,
  title={{WordNet}: a lexical database for English},
  author={Miller, George A},
  journal={Communications of the ACM},
  year={1995},
}

@article{wahcub200,
  title = {Caltech-UCSD Birds-200-2011},
  author = {Wah, C. and Branson, S. and Welinder, P. and Perona, P. and Belongie, S.},
  year = {2011},
  institution = {California Institute of Technology},
}

@inproceedings{khosla2011novel,
  title={Novel dataset for fine-grained image categorization: Stanford dogs},
  author={Khosla, Aditya and Jayadevaprakash, Nityananda and Yao, Bangpeng and Li, Fei-Fei},
  booktitle={CVPR workshop},
  year={2011},
}

@article{hu2021lora,
  title={{LoRA}: Low-rank adaptation of large language models},
  author={Hu, Edward J and Shen, Yelong and Wallis, Phillip and Allen-Zhu, Zeyuan and Li, Yuanzhi and Wang, Shean and Wang, Lu and Chen, Weizhu},
  journal={arXiv preprint arXiv:2106.09685},
  year={2021}
}

@InProceedings{Dong_2023_CVPR,
    author    = {Dong, Xiaoyi and Bao, Jianmin and Zheng, Yinglin and Zhang, Ting and Chen, Dongdong and Yang, Hao and Zeng, Ming and Zhang, Weiming and Yuan, Lu and Chen, Dong and Wen, Fang and Yu, Nenghai},
    title     = {MaskCLIP: Masked Self-Distillation Advances Contrastive Language-Image Pretraining},
    booktitle = {Proceedings of the IEEE/CVF Conference on Computer Vision and Pattern Recognition (CVPR)},
    month     = {June},
    year      = {2023},
    pages     = {10995-11005}
}

@InProceedings{Li_2023_CVPR,
    author    = {Li, Yanghao and Fan, Haoqi and Hu, Ronghang and Feichtenhofer, Christoph and He, Kaiming},
    title     = {Scaling Language-Image Pre-Training via Masking},
    booktitle = {Proceedings of the IEEE/CVF Conference on Computer Vision and Pattern Recognition (CVPR)},
    month     = {June},
    year      = {2023},
    pages     = {23390-23400}
}

@article{EVA-CLIP,
  title={Eva-clip: Improved training techniques for clip at scale},
  author={Sun, Quan and Fang, Yuxin and Wu, Ledell and Wang, Xinlong and Cao, Yue},
  journal={arXiv preprint arXiv:2303.15389},
  year={2023}
}

@inproceedings{maskQCLIP,
  title={MasQCLIP for Open-Vocabulary Universal Image Segmentation},
  author={Xu, Xin and Xiong, Tianyi and Ding, Zheng and Tu, Zhuowen},
  booktitle={Proceedings of the IEEE/CVF International Conference on Computer Vision},
  pages={887--898},
  year={2023}
}

@inproceedings{zang2022open,
  title={Open-vocabulary detr with conditional matching},
  author={Zang, Yuhang and Li, Wei and Zhou, Kaiyang and Huang, Chen and Loy, Chen Change},
  booktitle={ECCV},
  year={2022},
}

@inproceedings{ReCLIP,
  title={ReCLIP: A Strong Zero-Shot Baseline for Referring Expression Comprehension},
  author={Subramanian, Sanjay and Merrill, William and Darrell, Trevor and Gardner, Matt and Singh, Sameer and Rohrbach, Anna},
  booktitle={Proceedings of the 60th Annual Meeting of the Association for Computational Linguistics (Volume 1: Long Papers)},
  pages={5198--5215},
  year={2022}
}

@article{circleCLIP,
  title={What does clip know about a red circle? visual prompt engineering for vlms},
  author={Shtedritski, Aleksandar and Rupprecht, Christian and Vedaldi, Andrea},
  journal={arXiv preprint arXiv:2304.06712},
  year={2023}
}

@inproceedings{MaskAdaptedCLIP,
  title={Open-vocabulary semantic segmentation with mask-adapted clip},
  author={Liang, Feng and Wu, Bichen and Dai, Xiaoliang and Li, Kunpeng and Zhao, Yinan and Zhang, Hang and Zhang, Peizhao and Vajda, Peter and Marculescu, Diana},
  booktitle={CVPR},
  year={2023}
}

@InProceedings{clipseg,
    author    = {L\"uddecke, Timo and Ecker, Alexander},
    title     = {Image Segmentation Using Text and Image Prompts},
    booktitle = {Proceedings of the IEEE/CVF Conference on Computer Vision and Pattern Recognition (CVPR)},
    month     = {June},
    year      = {2022},
    pages     = {7086-7096}
}

@misc{fgvp,
      title={Fine-Grained Visual Prompting}, 
      author={Lingfeng Yang and Yueze Wang and Xiang Li and Xinlong Wang and Jian Yang},
      year={2023},
      eprint={2306.04356},
      archivePrefix={arXiv},
      primaryClass={cs.CV}
}

@article{glass2022re2g,
  title={Re2G: Retrieve, rerank, generate},
  author={Glass, Michael and Rossiello, Gaetano and Chowdhury, Md Faisal Mahbub and Naik, Ankita Rajaram and Cai, Pengshan and Gliozzo, Alfio},
  journal={arXiv preprint arXiv:2207.06300},
  year={2022}
}

@article{Yu2024RankRAGUC,
  title={RankRAG: Unifying Context Ranking with Retrieval-Augmented Generation in LLMs},
  author={Yue Yu and Wei Ping and Zihan Liu and Boxin Wang and Jiaxuan You and Chao Zhang and Mohammad Shoeybi and Bryan Catanzaro},
  journal={ArXiv},
  year={2024}
}

@article{liu2023dynamic,
  title={Dynamic multimodal fusion via meta-learning towards micro-video recommendation},
  author={Liu, Han and Wei, Yinwei and Liu, Fan and Wang, Wenjie and Nie, Liqiang and Chua, Tat-Seng},
  journal={ACM Transactions on Information Systems},
  volume={42},
  number={2},
  pages={1--26},
  year={2023},
  publisher={ACM New York, NY, USA}
}

@article{liu2022hs,
  title={HS-GCN: Hamming spatial graph convolutional networks for recommendation},
  author={Liu, Han and Wei, Yinwei and Yin, Jianhua and Nie, Liqiang},
  journal={IEEE Transactions on Knowledge and Data Engineering},
  volume={35},
  number={6},
  pages={5977--5990},
  year={2022},
  publisher={IEEE}
}

@article{bai2025qwen2,
  title={Qwen2. 5-vl technical report},
  author={Bai, Shuai and Chen, Keqin and Liu, Xuejing and Wang, Jialin and Ge, Wenbin and Song, Sibo and Dang, Kai and Wang, Peng and Wang, Shijie and Tang, Jun and others},
  journal={arXiv preprint arXiv:2502.13923},
  year={2025}
}

@software{ilharco_gabriel_2021_5143773,
  author       = {Ilharco, Gabriel and
                  Wortsman, Mitchell and
                  Wightman, Ross and
                  Gordon, Cade and
                  Carlini, Nicholas and
                  Taori, Rohan and
                  Dave, Achal and
                  Shankar, Vaishaal and
                  Namkoong, Hongseok and
                  Miller, John and
                  Hajishirzi, Hannaneh and
                  Farhadi, Ali and
                  Schmidt, Ludwig},
  title        = {OpenCLIP},
  month        = jul,
  year         = 2021,
  note         = {If you use this software, please cite it as below.},
  publisher    = {Zenodo},
  version      = {0.1},
  doi          = {10.5281/zenodo.5143773},
  url          = {https://doi.org/10.5281/zenodo.5143773}
}

@article{He2015DeepRL,
  title={Deep Residual Learning for Image Recognition},
  author={Kaiming He and X. Zhang and Shaoqing Ren and Jian Sun},
  journal={2016 IEEE Conference on Computer Vision and Pattern Recognition (CVPR)},
  year={2015},
  pages={770-778},
  url={https://api.semanticscholar.org/CorpusID:206594692}
}

@article{douze2024faiss,
      title={The Faiss library},
      author={Matthijs Douze and Alexandr Guzhva and Chengqi Deng and Jeff Johnson and Gergely Szilvasy and Pierre-Emmanuel Mazaré and Maria Lomeli and Lucas Hosseini and Hervé Jégou},
      year={2024},
      eprint={2401.08281},
      archivePrefix={arXiv},
      primaryClass={cs.LG}
}

@inproceedings{Detic,
  title={Detecting Twenty-thousand Classes using Image-level Supervision},
  author={Zhou, Xingyi and Girdhar, Rohit and Joulin, Armand and Kr{\"a}henb{\"u}hl, Philipp and Misra, Ishan},
  booktitle={ECCV},
  year={2022}
}


 




\newpage

\vfill

\end{document}